\documentclass{article}

\usepackage{iclr2027_conference,times}
\iclrfinalcopy % Reveal authors and disable the ICLR review line-number ruler.

\usepackage[utf8]{inputenc}
\DeclareUnicodeCharacter{FF5E}{\nobreakspace}
\usepackage{amsmath,amssymb}
\usepackage{graphicx}
\graphicspath{{figures/}}
\usepackage{booktabs}
\usepackage{longtable}
\usepackage{multirow}
\usepackage{array}
\usepackage{tabularx}
\usepackage[table]{xcolor}
\usepackage{microtype}
\usepackage{url}
\usepackage{xspace}
\usepackage{placeins}
\usepackage{float}
\usepackage{tikz}
\usetikzlibrary{arrows.meta,calc,fit,positioning}
\usepackage{pgfplots}
\usepgfplotslibrary{groupplots}
\pgfplotsset{compat=1.18}
\usepackage{hyperref}
\hypersetup{
  hidelinks,
  pdftitle={What Visual Representations Do World Action Models Need?},
  pdfauthor={Kai Liu, Jiangran Lyu, Tingyu Deng, Yixin Zheng, Jingtian Wan, Wenbo Cui, Mi Yan, Yizhou Zhou, Haoran Li, Dongbin Zhao, Zhizheng Zhang, He Wang},
  pdfsubject={Visual representations for world action models},
  pdfkeywords={world action models, visual representation learning, robot manipulation, systematic study}
}

\definecolor{baselinegray}{RGB}{105,105,105}
\definecolor{capappearance}{RGB}{49,130,117}
\definecolor{capgeometry}{RGB}{66,103,178}
\definecolor{capsemantic}{RGB}{221,132,47}
\definecolor{captemporal}{RGB}{139,89,164}

\title{What Visual Representations Do World Action Models Need?}

\author{%
  \begin{minipage}[t]{\dimexpr\textwidth-2\tabcolsep\relax}
  \centering\normalfont
  Kai Liu\textsuperscript{1,2,*}\quad
  Jiangran Lyu\textsuperscript{2,3,*}\quad
  Tingyu Deng\textsuperscript{3}\quad
  Yixin Zheng\textsuperscript{1,2}\\[3pt]
  Jingtian Wan\textsuperscript{2, 4}\quad
  Wenbo Cui\textsuperscript{1, 2}\quad
  Mi Yan\textsuperscript{2, 3}\quad
  Yizhou Zhou\textsuperscript{5}\\[3pt]
  Haoran Li\textsuperscript{1}\quad
  Dongbin Zhao\textsuperscript{1,\textdagger}\quad
  Zhizheng Zhang\textsuperscript{2,\textdagger}\quad
  He Wang\textsuperscript{2,3,\textdagger}\\[7pt]
  {\small
  \textsuperscript{1}Institute of Automation, Chinese Academy of Sciences\\
  \textsuperscript{2}Galbot\quad
  \textsuperscript{3}Peking University\\
  \textsuperscript{4}University of Science and Technology of China\\
  \textsuperscript{5}Individual Researcher\\[4pt]
  \textsuperscript{*}Equal contribution.\quad
  \textsuperscript{\textdagger}Co-corresponding authors.\\[6pt]
  \textbf{Project page:} \url{https://kailiu18.github.io/R-2WAM.github.io/}\endgraf}
  \end{minipage}%
}
\date{}

\begin{document}
\maketitle
% The ICLR style sets a publication header inside \maketitle. Override it
% afterwards: this is a preprint, with neither review nor acceptance claims.
\fancyhead{}
\renewcommand{\headrulewidth}{0pt}
\pagestyle{plain}
\thispagestyle{plain}

\begin{abstract}
World Action Models (WAMs) inherit powerful priors from pretrained video generation models, but their variational autoencoder (VAE) representations are optimized for visual reconstruction rather than manipulation. Recent studies show that external visual representations improve manipulation performance and robustness. However, these methods select representations based on presumed capability gaps and use different integration mechanisms, making it difficult to disentangle the effects of representation choice and integration. To better understand how external representations strengthen WAMs, we conduct a systematic study through controlled comparisons of multiple visual representations, five integration mechanisms, and feature combinations. Our study yields three main findings. First, strong general-purpose visual representations provide broader robustness gains and can match or outperform representations pretrained for specific visual capabilities. Second, channel-wise joint prediction achieves the highest average success rate; further analyses suggest that future-feature prediction and the action expert's effective use of added visual information contribute to its gains. Third, additional sources should be selected for the task-relevant information they contribute beyond existing representations, as the same source can help one combination but harm another. However, even the best evaluated composition provides a smaller additional gain than the initial improvement from adding a strong general-purpose representation to the VAE-only baseline. We instantiate the resulting \textbf{R}epresentation \textbf{R}ecipe as R\textsuperscript{2}-WAM. It achieves state-of-the-art performance on RoboCasa GR1 Tabletop with an 84.7\% success rate and reaches 92.8\% on RoboTwin~2.0, improving over the VAE-only baselines by 5.9 and 2.2 percentage points, respectively. Taken together, these findings suggest that future work should prioritize stronger general-purpose visual representations and more effective mechanisms for integrating them into WAMs, with feature composition serving as a supplementary route to improvement.
\end{abstract}
\section{Introduction}

World Action Models (WAMs) adapt pretrained video generation models to robot control by jointly predicting future video and actions. This allows robot policies to draw on the visual structure and temporal dynamics learned through large-scale video pretraining, as demonstrated by DreamZero~\citep{dreamzero}, LingBot-VA~\citep{lingbotva}, and Fast-WAM~\citep{fastwam}. Along with these priors, WAMs inherit the latent space of their variational autoencoders (VAEs).

% The VAE is trained to reconstruct visual observations, including details that may be irrelevant to manipulation. Its reconstruction objective is not explicitly designed for robot manipulation. Whether the resulting representation adequately supports the perceptual demands of manipulation, such as semantic understanding and geometric reasoning, remains an empirical question.

The inherited VAE is optimized for visual reconstruction~\citep{wan}, but how well its representations support robot manipulation remains unclear. Its reconstruction objective does not explicitly prioritize manipulation-relevant information, such as object semantics and spatial relationships. Recent studies augment WAMs with external visual features and report improvements in manipulation performance and robustness~\citep{agra,robustwam,geosemwam}. However, these methods select different representations based on prior assumptions about which visual capabilities WAMs lack. Systematic comparisons across representation types remain limited, leaving unclear which visual representations best support robot manipulation in WAMs.

% How to integrate external representations effectively, and whether combining more sources yields continued gains, also remain open questions for WAMs. 

Beyond representation choice, WAM performance also depends on the mechanism used to integrate external features into the model~\citep{fourdwam}. A further question is whether combining more representations leads to continued performance gains. External features can be integrated through representation alignment~\citep{repa,agra}, action conditioning~\citep{stwam}, or joint prediction~\citep{flexpi}. Existing studies often couple selected representations with specific integration designs~\citep{agra,flexpi,mecowam}, making the effects of representation choice and integration mechanism difficult to disentangle across methods. Besides, the diverse demands of manipulation---including semantics, geometry, object extent, motion, and interaction---suggest that different representations may be needed. Yet additional sources may also introduce redundant information or interference during optimization. Whether and when different feature combinations improve WAM performance therefore remains an empirical question.

To address these questions, we systematically study visual representations in WAMs, focusing on capabilities, integration mechanisms, and composition. Using a common WAM backbone built on a pretrained video generation model, we compare diverse visual representations, five integration mechanisms, and multiple source combinations under a shared training protocol. Our study addresses three research questions (RQs):
\begin{description}
    \setlength{\itemsep}{0pt}
    \setlength{\parsep}{0pt}
    \item[RQ1---Representations:] Which visual representations improve WAM performance?
    \item[RQ2---Mechanisms:] Which feature injection mechanism yields the greatest performance gains from external features?
    \item[RQ3---Composition:] Does combining more feature sources consistently improve WAM performance?
\end{description}

\begin{figure}[t]
\centering
\begin{tikzpicture}
\node[anchor=south west,inner sep=0] (teaser) at (0,0) {\includegraphics[width=\textwidth]{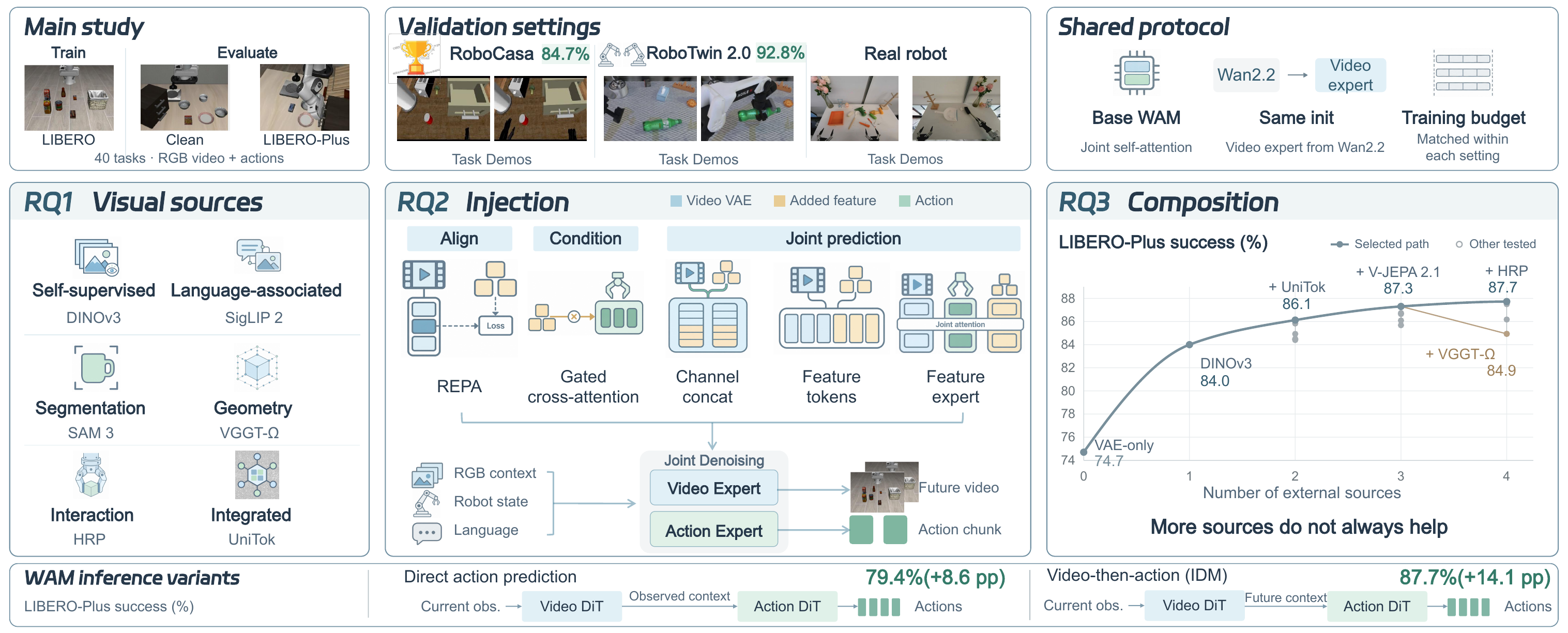}};
\end{tikzpicture}
\caption{Overview of our systematic study of visual representations in WAMs. Under a shared training protocol, we compare visual feature sources (RQ1), five integration mechanisms (RQ2), and feature compositions (RQ3). }
\label{fig:teaser}
\end{figure}

% Our main experiments combine simulation evaluations on LIBERO and LIBERO-Plus with real-robot experiments on a Galbot dual-arm platform. We further evaluate the resulting representation design on RoboCasa and RoboTwin 2.0.

Our controlled comparisons yield three findings that guide the selection, integration, and composition of external visual representations for WAMs. \textbf{(1) General-purpose visual representations should be the default choice for augmenting WAMs.} They can match or outperform representations pretrained for specific visual capabilities---even under perturbations relevant to those capabilities---while providing broader robustness gains. \textbf{(2) Channel-wise joint prediction yields the largest average gains across matched sources.} Our analyses suggest that future-feature prediction and effective use of the added visual information by the action expert contribute to these gains and should guide future integration designs. \textbf{(3) Additional sources should be selected for the task-relevant information they add beyond the existing representations.} The same source can improve one combination but provide little benefit or even reduce performance in another. Although feature composition can provide further gains, these are limited compared with the initial improvement. Feature composition should therefore serve as a supplementary route to improvement.

% We instantiate these findings as a \textbf{R}epresentation \textbf{R}ecipe for WAMs, termed R\textsuperscript{2}-WAM, and evaluate its effectiveness across WAM formulations and benchmarks. On LIBERO-Plus, R\textsuperscript{2}-WAM achieves success rates of 81.66\% under direct action prediction and 87.72\% under video-then-action prediction, improving over the corresponding VAE-only baselines by $[\Delta_1]$ and $[\Delta_2]$ percentage points, respectively. It achieves state-of-the-art performance on RoboCasa GR1 Tabletop with an 84.4\% success rate and reaches 92.9\% on RoboTwin~2.0, exceeding the corresponding VAE-only baselines by 5.6 and 2.3 percentage points, respectively. These results support the applicability of our representation recipe beyond the WAM formulation and benchmarks used in the controlled study.

Our contributions are threefold:
\begin{itemize}
    \setlength{\itemsep}{0pt}
    \setlength{\parsep}{0pt}
    \item \textbf{A controlled study of visual representations in WAMs.} We systematically compare representation sources, five integration mechanisms, and feature combinations within a common WAM backbone and shared training protocol, enabling controlled comparisons of their effects on manipulation performance and robustness.
    \item \textbf{Empirical guidance for WAM representation design.} We derive practical recommendations for selecting, integrating, and combining external visual representations. These findings motivate future research on stronger general-purpose visual foundation models and more efficient and effective mechanisms for integrating their representations into WAMs.
    \item \textbf{R\textsuperscript{2}-WAM and validation beyond the controlled study.} We instantiate these findings as a \textbf{R}epresentation \textbf{R}ecipe for WAMs, termed R\textsuperscript{2}-WAM. We validate the effectiveness of the recipe across different WAM formulations on LIBERO-Plus, and further demonstrate improvements over the corresponding VAE-only baselines on RoboCasa GR1 Tabletop and RoboTwin~2.0, as well as in real-world robot experiments.
\end{itemize}

\section{Related Work}

\paragraph{World Action Models.}
WAMs learn action prediction alongside future visual dynamics. DreamZero~\citep{dreamzero} transfers a pretrained video diffusion backbone to joint video--action generation, and LingBot-VA~\citep{lingbotva} combines autoregressive diffusion with closed-loop visual feedback. Fast-WAM~\citep{fastwam} shows that video co-training can benefit control even when future video generation is omitted at inference. Other studies modify the visual prediction space or its supervision. LDA-1B~\citep{lda1b} learns dynamics in DINO feature space, while LaWAM~\citep{lawam} conditions action generation on predicted latent visual subgoals. RepWAM~\citep{repwam} and LingBot-VA~2.0~\citep{lingbotva2} learn semantically aligned visual--action tokenizers. ST-WAM~\citep{stwam} diagnoses failures under visual shifts and combines DINO future prediction and history retrieval with VAE dynamics. Robust-WAM~\citep{robustwam} aligns queries in the action stream with future semantic features while retaining the VAE generation pathway. Flex-$\pi$~\citep{flexpi} jointly denoises RGB latents, 3D pointmaps, DINO features, and actions, supporting different combinations of input and prediction streams at deployment. These studies demonstrate the effectiveness of specific visual representations within their respective WAM designs, but systematic comparisons are still needed to identify which features complement the capabilities that WAMs inherit from pretrained video generators.

\paragraph{Visual Representations.}
Pretraining objectives shape the information captured by visual representations. Visual self-supervision has produced encoders such as DINOv3~\citep{dinov3}, VC-1~\citep{vc1}, LingBot-Vision~\citep{lingbotvision}, VideoMAE~V2~\citep{videomaev2}, and V-JEPA~2.1~\citep{vjepa21}. Language-associated approaches include image--text models such as CLIP~\citep{clip} and SigLIP~2~\citep{siglip2}, as well as multimodal language models such as Qwen3.5~\citep{qwen35}. Other pretraining objectives target segmentation with SAM~3~\citep{sam3}, geometry with VGGT-$\Omega$~\citep{vggtomega}, and interaction-related information with HRP~\citep{hrp} and DynaFLIP~\citep{dynaflip}. UniTok~\citep{unitok} and C-RADIOv4~\citep{cradiov4} further combine multiple learning objectives or teacher models within a single representation. These approaches provide different but overlapping visual capabilities, but how their pretraining emphases translate into manipulation gains in WAMs remains an empirical question.

\section{Study Design}

\subsection{Base WAM Architecture}
\label{sec:study-setting}

Given RGB observations, a language instruction, and the current robot state, the base WAM jointly predicts future-video latents $z_{1:T}^{\mathrm{vae}}$ and an action chunk $a$. The model comprises a video expert initialized from Wan~2.2~\citep{wan} and a separate action expert. The experts interact through joint attention with an
asymmetric mask: action tokens attend to both video and
action tokens, whereas video tokens attend only to video
tokens. Both experts attend to T5-encoded language features~\citep{t5}
through cross-attention and are additionally conditioned
on the robot state and flow time.

Throughout, $t=0$ denotes the current observation, $t=1,\ldots,T$ indexes the future sequence, and $\tau$ denotes flow time. Thus $z_0^{\mathrm{vae}}$ denotes current-frame VAE latents, and $z_{1:T,\tau}^{\mathrm{vae}}$ denotes future-video latents at flow time $\tau$.

We train the model with a weighted sum of video and action
flow-matching losses~\citep{flowmatching}:
\begin{equation}
\mathcal L_{\mathrm{WAM}}
=
\mathcal L_{\mathrm{FM}}(z_{1:T}^{\mathrm{vae}})
+
\lambda_{\mathrm{act}}\mathcal L_{\mathrm{FM}}(a),
\end{equation}
where $\mathcal L_{\mathrm{FM}}$ denotes the modality-specific
mean-squared error between predicted and target flow
velocities, and $\lambda_{\mathrm{act}}$ weights the action loss.
At inference, future-video latents and actions are initialized
from noise and jointly denoised.

\subsection{Visual Representations}
\label{sec:feature-sources}

We compare thirteen pretrained visual encoders, using their pretraining objectives as an initial indication of the information their representations capture and the visual capabilities they may support. All encoders remain frozen during WAM training and extract representations from the RGB observations available to the base WAM. We consider six pretraining emphases:

\textbf{Visual self-supervision:} VC-1~\citep{vc1} and VideoMAE~V2~\citep{videomaev2} learn visual structure through masked image or video reconstruction. V-JEPA~2.1~\citep{vjepa21} predicts visual features to learn spatial and temporal relationships; DINOv3~\citep{dinov3} uses self-distillation to learn semantic and dense spatial features; and LingBot-Vision~\citep{lingbotvision} emphasizes fine spatial structure through masked boundary modeling.

\textbf{Language-associated pretraining:} SigLIP~2~\citep{siglip2} combines image--text alignment with captioning and self-supervised objectives to learn semantic and dense visual features. Qwen3.5~\citep{qwen35} uses multimodal language modeling for diverse visual understanding tasks.

\textbf{Segmentation supervision:} SAM~3~\citep{sam3} learns object regions and boundaries through concept-conditioned detection and segmentation.

\textbf{Geometric supervision:} VGGT-$\Omega$~\citep{vggtomega} learns scene geometry through 3D reconstruction.

\textbf{Interaction-oriented pretraining:} HRP~\citep{hrp} uses hand, object, and contact supervision to learn manipulation affordances. DynaFLIP~\citep{dynaflip} aligns images, language, and 3D motion to capture information relevant to interaction dynamics.

\textbf{Multi-objective learning and multi-teacher distillation:} UniTok~\citep{unitok} combines reconstruction and image--text learning to retain visual detail and semantics, while C-RADIOv4~\citep{cradiov4} distills DINOv3, SigLIP 2, and SAM 3.

Based on these pretraining objectives and capability emphases, we designate representations intended for broad transfer across visual or vision--language tasks as \emph{general-purpose}, and those pretrained primarily for specific visual capabilities as \emph{specialized}. Table~\ref{tab:representation-capabilities} lists the sources and their group assignments.

\subsection{Feature Integration Mechanisms}
\label{sec:feature-injection}
\label{sec:multi-teacher}

\begin{figure}[t]
\centering
\includegraphics[width=\textwidth]{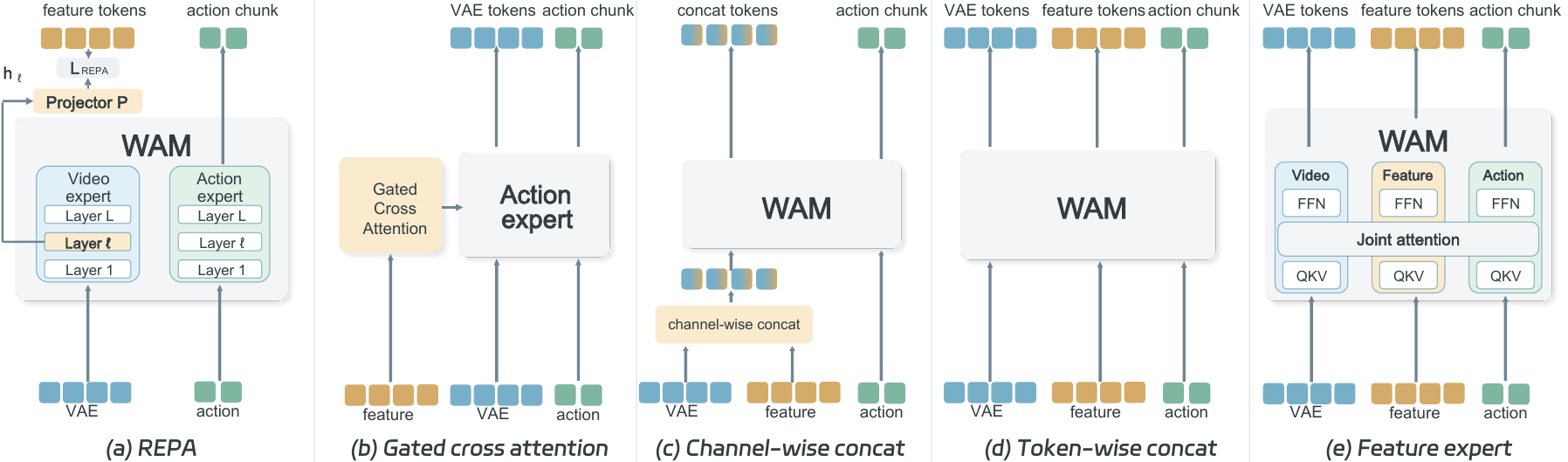}
\caption{Injection mechanisms: (a) REPA, (b) gated cross attention, (c) channel-wise concat, (d) token-wise concat, and (e) feature expert.}
\label{fig:pipeline}
\end{figure}

We compare five mechanisms that use external features
for representation alignment, action conditioning, or
joint prediction (Figure~\ref{fig:pipeline}). The three joint-prediction mechanisms
differ in how feature and VAE representations are fused
and whether their Transformer parameters are shared.

We denote current and future external features by $f_0$ and $f_{1:T}$, respectively, with a superscript $k$ identifying the source when needed. The state of future features at flow time $\tau$ is written as $f_{1:T,\tau}$.

\textbf{Representation alignment (REPA):}
A trainable projector maps intermediate WAM hidden states into the external feature space. A cosine-distance loss aligns the projected states with spatially corresponding features from the frozen source encoder~\citep{repa}. 

\textbf{Gated cross-attention:}
An additional cross-attention module uses action hidden states as queries and projected features from the current observations as keys and values. A learnable gate scales the module's output before it is added to the action hidden states~\citep{flamingo}. The existing cross-attention to T5 language features is retained. 

\textbf{Channel-wise concatenation:}
We spatially align external features with the VAE latent
grid and concatenate them along the channel dimension.
The resulting representation is processed by the video
expert, whose input and output projections are expanded
to jointly predict VAE and external-feature flow velocities.
The video-token layout and attention mask remain
unchanged~\citep{dreamworld}.

\textbf{Token-wise concatenation.}
We represent external features as a separate token sequence
with its own input projection and future-feature prediction
head. Feature and VAE tokens share the video expert's
Transformer layers~\citep{flexpi}.

\textbf{Feature expert:}
We introduce a dedicated Transformer expert to predict
future external features. Unlike token-wise concatenation,
this mechanism uses independent feature-processing
parameters. Feature, video, and action tokens interact
through masked joint attention, while each expert retains
its own projections and feed-forward layers~\citep{motus}.

For REPA and the three joint-prediction mechanisms, the training objective is
\begin{equation}
\mathcal L=\mathcal L_{\mathrm{WAM}}+\lambda_{\mathrm{aux}}\mathcal L_{\mathrm{aux}},
\label{eq:objective}
\end{equation}
where $\mathcal L_{\mathrm{aux}}$ is the cosine-distance alignment loss for REPA and the future-feature flow-matching loss for the joint-prediction mechanisms. Gated cross-attention is trained using $\mathcal L_{\mathrm{WAM}}$ alone.

We evaluate feature composition using channel-wise concatenation as the common integration mechanism. Features from each source are spatially aligned with the VAE latent grid and concatenated along the channel dimension.

\section{Experiments}
\label{sec:experiments}

\subsection{Experimental Setup}
\label{sec:study-design}

\paragraph{Training.}
We use the LIBERO～\citep{libero} and RoboTwin~2.0~\citep{robotwin2} training datasets from Fast-WAM~\citep{fastwam} and the RoboCasa GR1 Tabletop~\citep{robocasa} dataset used in DIAL~\citep{dial}. Across the LIBERO comparisons (RQ1--RQ3), we hold the training data, base observation inputs, action horizon, optimizer, and learning-rate schedule fixed while varying representation sources, integration mechanisms, or source combinations. Training configurations for simulation and real-robot experiments are provided in Appendix~\ref{app:training-configuration}.

\paragraph{Evaluation.}
We evaluate LIBERO~\citep{libero} with 50 rollouts per task following Fast-WAM~\citep{fastwam}; full results are reported in the Appendix~\ref{app:full-mechanism-results} since performance is near saturation. The same models are evaluated on LIBERO-Plus (LP)~\citep{liberoplus} without further training, using three evaluation seeds and all 10,030 episodes per seed across seven perturbation categories. We report per-category and overall success rates. RoboCasa GR1 Tabletop~\citep{robocasa,dial} uses 50 trials per task across 24 tasks, while RoboTwin~2.0~\citep{robotwin2,fastwam} uses 100 trials per task across 50 tasks in each of the clean and randomized settings. We evaluate the VAE-only WAM and R\textsuperscript{2}-WAM on four real-robot tasks with 20 trials per task per model; metric definitions are provided in Appendix~\ref{app:real-robot-evaluation}.

\subsection{RQ1: Which visual representations improve WAM performance?}
\label{sec:representation-study}

We compare the visual representations introduced in Section~\ref{sec:feature-sources}.
All representations are integrated using channel-wise concatenation, with the VAE-only WAM serving as the baseline.
Table~\ref{tab:representation-capabilities} reports success rates as mean $\pm$ standard deviation across three evaluation seeds for each of the seven LP perturbation categories and overall.
We examine both overall performance and the range of perturbations under which each source improves robustness.

\begin{table}[!t]
\centering
\caption{RQ1: single-source success rates (\%) across all seven LIBERO-Plus～\citep{liberoplus} perturbations under channel-wise concatenation. Sources are grouped as general-purpose or specialized, with subgroups indicating the dominant pretraining emphasis (Section~\ref{sec:feature-sources}). Results are reported as mean $\pm$ standard deviation across three evaluation seeds.}
\label{tab:representation-capabilities}
\begingroup
\normalsize
\setlength{\tabcolsep}{0.4pt}
\renewcommand{\arraystretch}{1.0}
\newcommand{\rqonestat}[2]{\mbox{#1\textpm#2}}
\newcommand{\rqonebest}[2]{\mbox{\textbf{#1\textpm#2}}}
\begin{tabular*}{\linewidth}{@{\extracolsep{\fill}}lcccccccc@{}}
\toprule
Source & Camera & Robot & Lang. & Light & Bkg. & Noise & Layout & Total \\
\midrule
VAE-only & \rqonestat{51.2}{0.6} & \rqonestat{77.8}{0.8} & \rqonestat{91.1}{1.2} & \rqonestat{94.3}{2.4} & \rqonestat{62.0}{1.4} & \rqonestat{66.6}{1.2} & \rqonestat{82.4}{0.2} & \rqonestat{74.7}{0.5} \\
\addlinespace[2pt]
\rowcolor{black!12}
\multicolumn{9}{@{}l}{\textbf{General-purpose representations}} \\
\rowcolor{black!4}
\multicolumn{9}{@{}l}{\emph{Visual self-supervision}} \\
DINOv3 & \rqonebest{64.2}{1.4} & \rqonestat{82.5}{0.9} & \rqonestat{93.8}{0.8} & \rqonebest{98.5}{0.4} & \rqonestat{83.9}{0.4} & \rqonestat{86.0}{0.3} & \rqonestat{83.6}{0.8} & \rqonebest{84.0}{0.4} \\
V-JEPA~2.1 & \rqonestat{63.0}{0.5} & \rqonestat{80.9}{0.8} & \rqonestat{90.9}{0.5} & \rqonestat{96.8}{0.5} & \rqonestat{80.7}{1.2} & \rqonestat{84.4}{0.5} & \rqonebest{84.2}{0.5} & \rqonestat{82.4}{0.4} \\
VC-1 & \rqonestat{60.8}{1.0} & \rqonestat{78.1}{0.5} & \rqonestat{93.3}{0.4} & \rqonestat{98.4}{0.4} & \rqonestat{82.0}{0.6} & \rqonestat{86.7}{1.3} & \rqonestat{83.0}{0.9} & \rqonestat{82.5}{0.4} \\
LingBot-Vision & \rqonestat{62.6}{1.9} & \rqonestat{79.6}{1.2} & \rqonestat{93.0}{0.4} & \rqonestat{97.8}{0.9} & \rqonestat{81.2}{0.3} & \rqonebest{87.7}{0.5} & \rqonestat{82.9}{0.7} & \rqonestat{83.0}{0.7} \\
VideoMAE V2 & \rqonestat{53.6}{1.4} & \rqonestat{81.4}{0.8} & \rqonestat{91.3}{1.1} & \rqonestat{95.4}{0.3} & \rqonestat{68.0}{1.0} & \rqonestat{73.1}{0.5} & \rqonestat{82.7}{0.5} & \rqonestat{77.5}{0.6} \\
\addlinespace[1pt]
\rowcolor{black!4}
\multicolumn{9}{@{}l}{\emph{Language-associated pretraining}} \\
SigLIP~2 & \rqonestat{60.7}{1.0} & \rqonestat{85.0}{0.7} & \rqonestat{93.8}{0.5} & \rqonestat{97.9}{0.7} & \rqonestat{82.4}{1.1} & \rqonestat{83.4}{0.3} & \rqonestat{83.0}{0.7} & \rqonestat{83.1}{0.4} \\
Qwen3.5 & \rqonestat{59.7}{2.6} & \rqonestat{84.5}{0.1} & \rqonestat{93.2}{0.2} & \rqonestat{98.0}{0.2} & \rqonestat{80.0}{1.3} & \rqonestat{78.7}{0.6} & \rqonestat{82.5}{1.1} & \rqonestat{81.6}{0.7} \\
\addlinespace[1pt]
\rowcolor{black!4}
\multicolumn{9}{@{}l}{\emph{Multi-objective / multi-teacher learning}} \\
UniTok & \rqonestat{63.0}{1.4} & \rqonebest{85.2}{0.6} & \rqonebest{93.9}{1.0} & \rqonestat{96.7}{0.9} & \rqonestat{76.6}{0.5} & \rqonestat{77.1}{0.8} & \rqonestat{83.9}{0.6} & \rqonestat{81.9}{0.4} \\
C-RADIOv4 & \rqonestat{62.6}{1.5} & \rqonestat{81.0}{1.3} & \rqonestat{92.2}{1.0} & \rqonestat{97.3}{1.2} & \rqonebest{86.2}{0.8} & \rqonestat{86.6}{0.7} & \rqonestat{83.6}{0.8} & \rqonestat{83.5}{0.6} \\
\addlinespace[2pt]
\rowcolor{black!12}
\multicolumn{9}{@{}l}{\textbf{Specialized representations}} \\
\rowcolor{black!4}
\multicolumn{9}{@{}l}{\emph{Segmentation supervision}} \\
SAM~3 & \rqonestat{57.8}{1.3} & \rqonestat{80.8}{1.5} & \rqonestat{93.7}{0.7} & \rqonestat{95.6}{1.2} & \rqonestat{75.3}{1.0} & \rqonestat{81.5}{1.0} & \rqonestat{82.4}{0.6} & \rqonestat{80.6}{0.4} \\
\addlinespace[1pt]
\rowcolor{black!4}
\multicolumn{9}{@{}l}{\emph{Geometric supervision}} \\
VGGT-$\Omega$ & \rqonestat{62.4}{0.7} & \rqonestat{78.8}{0.8} & \rqonestat{92.7}{0.4} & \rqonestat{95.8}{0.6} & \rqonestat{75.7}{0.2} & \rqonestat{79.6}{0.5} & \rqonestat{82.5}{0.9} & \rqonestat{80.6}{0.3} \\
\addlinespace[1pt]
\rowcolor{black!4}
\multicolumn{9}{@{}l}{\emph{Interaction-oriented pretraining}} \\
HRP & \rqonestat{52.6}{1.0} & \rqonestat{77.8}{1.0} & \rqonestat{91.9}{0.3} & \rqonestat{96.9}{1.0} & \rqonestat{82.2}{1.3} & \rqonestat{81.2}{0.8} & \rqonestat{82.4}{0.6} & \rqonestat{79.8}{0.4} \\
DynaFLIP & \rqonestat{55.2}{0.9} & \rqonestat{83.1}{1.0} & \rqonestat{92.3}{0.6} & \rqonestat{93.9}{1.4} & \rqonestat{69.9}{0.1} & \rqonestat{77.5}{0.8} & \rqonestat{82.1}{1.3} & \rqonestat{78.8}{0.3} \\
\bottomrule
\end{tabular*}
\endgroup
\par\vspace{3pt}
\end{table}~

\paragraph{Viewpoint, background, and noise perturbations expose key robustness gaps in the VAE-only WAM.}
The VAE-only WAM achieves high success rates under language perturbations ($91.1\%$) and lighting changes ($94.3\%$), yet performs substantially worse under viewpoint changes ($51.2\%$), background changes ($62.0\%$), and image noise ($66.6\%$).
This uneven robustness motivates examining whether external visual representations can address these weaknesses and improve performance across perturbation categories.

\paragraph{General-purpose representations show stronger robustness across perturbations.}
Across the sources evaluated in Table~\ref{tab:representation-capabilities}, general-purpose representations achieve a higher average success rate than specialized representations ($82.2\%$ vs. $80.0\%$).
A general-purpose source also achieves the highest observed mean success rate in each of the seven perturbation categories, including camera, background, and image noise, where the VAE-only WAM is weakest.

\paragraph{Takeaway 1. General-purpose visual representations should be the default for augmenting WAMs.}
Rather than choosing a specialized encoder to remedy an assumed missing capability, start with a general-purpose visual representation.
In our study, general-purpose representations provide broader robustness gains and can match or exceed specialized alternatives even where their specialization appears relevant.

\subsection{RQ2: Which feature integration mechanism yields the greatest performance gains from external features?}
\label{sec:injection-study}

We compare the five integration mechanisms in Section~\ref{sec:feature-injection} across thirteen visual representations (Figure~\ref{fig:rq2_mechanism_analysis}(a); per-source results in Appendix~\ref{app:full-mechanism-results}). We then examine future-feature prediction, predicted feature structure, and action sensitivity to investigate the gains from channel-wise concatenation (Figure~\ref{fig:rq2_mechanism_analysis}(b--d)).

\begin{figure}[t]
\centering
\includegraphics[width=\linewidth]{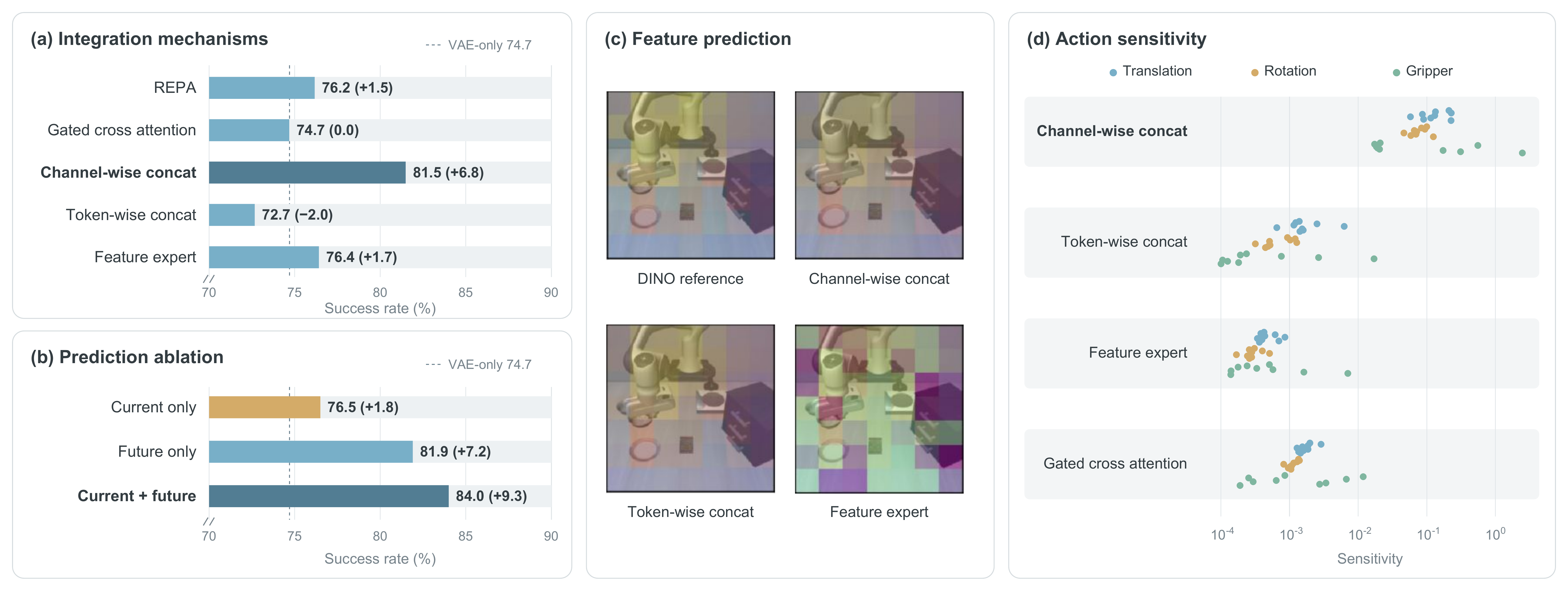}
\caption{\textbf{Feature-integration performance and mechanism analysis.}
(a) Average LP success rates across the thirteen representations in Table~\ref{tab:representation-capabilities}. (b) DINOv3～\citep{dinov3} prediction ablations within channel-wise concat.
(c) Shared-PCA features overlaid on future RGB observations. 
(d) Local sensitivity of final normalized actions to current DINOv3 features, inspired by gradient-based sensitivity and attribution~\citep{simonyan2013saliency,hoffman2019jacobian,yu2023sensory}. We evaluate action sensitivity at three selected time steps in each trajectory, corresponding to approach, grasp, and placement. Each point reports the sensitivity of one action group in the predicted action chunk at a selected time step.  Metric and computation details are in Appendix~\ref{app:action-sensitivity}.}
\label{fig:rq2_mechanism_analysis}
\end{figure}

\paragraph{Channel-wise concatenation achieves the highest average success rate.}
Across the thirteen visual representations in Table~\ref{tab:representation-capabilities}, channel-wise concatenation achieves an average LP success rate of $81.5\%$, exceeding the next-best mechanism by 5.1 percentage points (Figure~\ref{fig:rq2_mechanism_analysis}(a)).

\paragraph{Why channel-wise concatenation performs well.}

Figure~\ref{fig:rq2_mechanism_analysis}(b) separates current-feature conditioning from future-feature prediction. Future-only prediction reaches 81.9\%, close to the full configuration (84.0\%) and above current-only conditioning (76.5\%), supporting future-feature prediction as a major contributor.

The shared-PCA visualizations in Figure~\ref{fig:rq2_mechanism_analysis}(c) show that the video expert recovers coarse spatial structure in future features. Token-wise predictions also retain recognizable structure, whereas the feature-expert prediction is coarser in the displayed example.

We analyze how action generation depends on external features (Figure~\ref{fig:rq2_mechanism_analysis}(d)). Channel-wise concatenation exhibits higher translation and rotation sensitivities in these states, indicating stronger dependence of action generation on the added visual representations.

\paragraph{Takeaway 2. Use channel-wise concatenation as the default.}
For new designs, support future-feature prediction and encourage the action expert to make effective use of the added features.
Check prediction quality and how added features affect generated actions, and verify that design changes improve task success rates.

\subsection{RQ3: Does combining more feature sources consistently improve WAM performance?}
\label{sec:composition-study}

\begin{figure}[t]
\centering
\includegraphics[width=\linewidth]{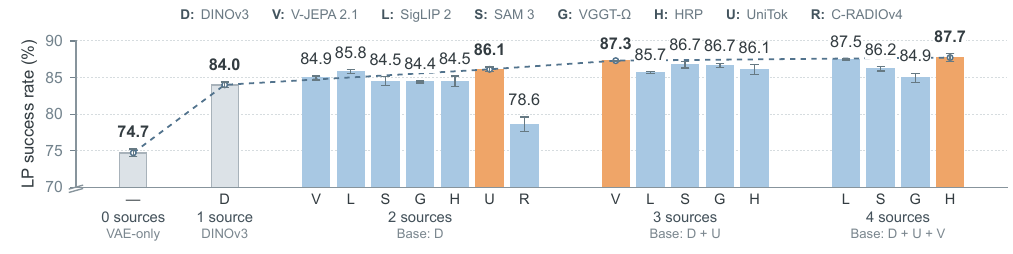}
\caption{RQ3: progressive feature composition on LIBERO-Plus (mean $\pm$ SD over three evaluation seeds). Each bar in the two-, three-, and four-source groups adds its labeled source to the indicated base.}
\label{fig:progressive-composition}
\end{figure}

Building on RQ1 and RQ2, we use DINOv3~\citep{dinov3} as the initial external representation and keep channel-wise concatenation fixed.
The candidate additions span different pretraining emphases, with representatives from each category in Section~\ref{sec:feature-sources}: V-JEPA~2.1~\citep{vjepa21} for visual self-supervision, SigLIP~2~\citep{siglip2} for language-associated learning, SAM~3~\citep{sam3} for segmentation, VGGT-$\Omega$~\citep{vggtomega} for geometry, HRP~\citep{hrp} for interaction-oriented pretraining, and UniTok~\citep{unitok} and C-RADIOv4~\citep{cradiov4} for multi-objective / multi-teacher learning.
We first add each candidate separately to DINOv3, then evaluate selected larger combinations through single-source additions (Figure~\ref{fig:progressive-composition}).

\paragraph{A general-purpose source provides the main gain.}
DINOv3~\citep{dinov3} alone improves LP success by $9.3$ percentage points over VAE-only, while adding other features yields at most a further $3.7$-point gain among the evaluated combinations.

\paragraph{The benefit of an additional source depends on the existing representations.}
The composition results suggest that overlap in pretraining objectives can limit complementary gains.
Both UniTok~\citep{unitok} and SigLIP~2~\citep{siglip2} learn language-aligned visual representations.
Adding SigLIP~2 improves DINOv3~\citep{dinov3} by $1.8$ percentage points, but reduces success by $0.4$ points once UniTok is included and adds only $0.2$ points when V-JEPA~2.1~\citep{vjepa21} is also present.
Spatial capabilities similarly overlap: DINOv3 already provides dense features supporting segmentation and depth estimation, while VGGT-$\Omega$~\citep{vggtomega} uses a DINOv3-initialized visual encoder.
Adding VGGT-$\Omega$ or SAM~3~\citep{sam3} to DINOv3 produces little change, although both improve some larger combinations.
A direct teacher--student relationship occurs with C-RADIOv4~\citep{cradiov4}, which distills DINOv3, SigLIP~2, and SAM~3: combining it with DINOv3 reduces LP success to $78.6\%$.
Meanwhile, adding V-JEPA~2.1~\citep{vjepa21} to DINOv3 and UniTok improves performance, consistent with its video-based predictive pretraining providing complementary cues.

\paragraph{Takeaway 3. Additional sources should be selected for complementarity.}
Build on a strong general-purpose representation and use composition for selective refinement.
When choosing additional sources, consider which task-relevant information they may contribute beyond the existing features, accounting for overlap in pretraining objectives and teacher models.

\begin{figure}[b]
\centering
\includegraphics[width=\linewidth]{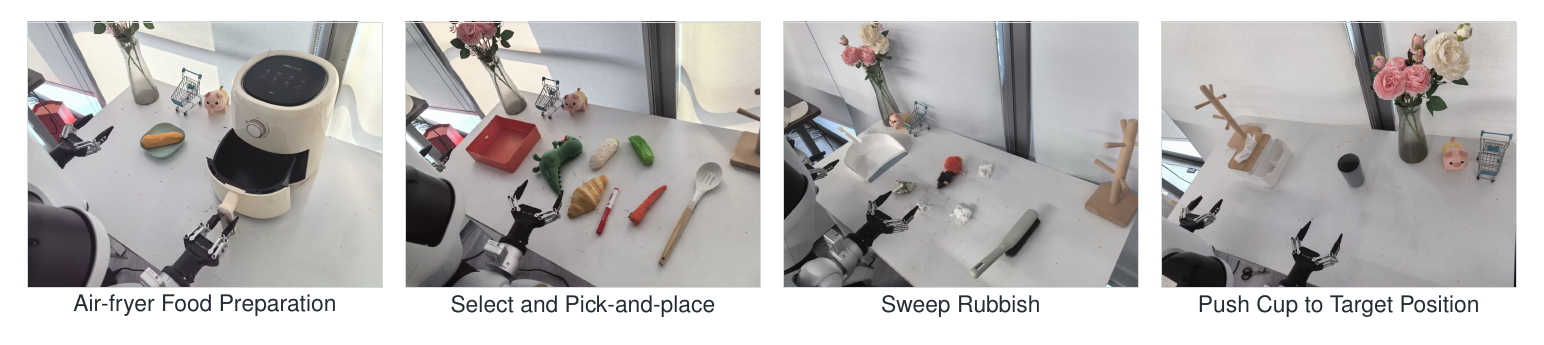}
\caption{Real-robot evaluation on four physical manipulation tasks.}
\label{fig:real-robot-tasks}
\end{figure}

\subsection{Representation Recipe and Validation}
\label{sec:generalization-study}

R\textsuperscript{2}-WAM combines DINOv3~\citep{dinov3}, UniTok~\citep{unitok}, and V-JEPA~2.1~\citep{vjepa21} through channel-wise concatenation. Adding HRP~\citep{hrp} yields the highest mean LP success ($87.7\%$ versus $87.3\%$), but only a $0.4$-percentage-point gain. We retain three sources to limit auxiliary encoders and feature-prediction targets.

\paragraph{Validation across WAM formulations.}
Table~\ref{tab:architecture-generalization}(a) compares VAE-only baselines with their recipe-augmented counterparts under two additional WAM formulations: direct action prediction~\citep{fastwam} and video-then-action prediction (IDM)~\citep{lingbotva}. Implementation details are provided in Appendix~\ref{app:wam-formulations}. These results support the benefit of the recipe across both evaluated formulations.

\begin{table}[t]
\centering
\caption{Validation Results across WAM formulations (a) and real-robot tasks (b). Real-robot evaluation uses 20 trials per task; metric definitions are in Appendix~\ref{app:real-robot-evaluation}.}
\label{tab:recipe-validation}
\label{tab:architecture-generalization}
\label{tab:r2wam-real-robot}
\begingroup
\normalsize
\setlength{\tabcolsep}{4pt}
\renewcommand{\arraystretch}{1.05}
\begin{minipage}[t]{0.47\linewidth}
\centering
\textbf{(a) WAM formulations}\par\smallskip
\begin{tabular}{@{}lrr@{}}
\toprule
Formulation & LIBERO & LP \\
\midrule
Direct action prediction & 96.8 & 70.8 \\
\quad + recipe & \textbf{98.4} & \textbf{79.4} \\
\midrule
Video-then-action (IDM) & 98.6 & 73.6 \\
\quad + recipe & \textbf{99.1} & \textbf{87.7} \\
\bottomrule
\end{tabular}
\end{minipage}\hfill
\begin{minipage}[t]{0.51\linewidth}
\centering
\textbf{(b) Real-robot tasks}\par\smallskip
\begin{tabular}{@{}lrr@{}}
\toprule
Task & VAE-only & R\textsuperscript{2}-WAM \\
\midrule
Air-fryer Food Preparation & 50.0 & \textbf{60.0} \\
Select and Pick-and-place & 70.0 & \textbf{80.0} \\
Sweep Rubbish & 48.3 & \textbf{66.7} \\
Push Cup to Target Position & 65.0 & \textbf{75.0} \\
\bottomrule
\end{tabular}
\end{minipage}
\endgroup
\end{table}

\paragraph{Cross-benchmark validation.}
\begin{table}[t]
\centering
\caption{Comparison with selected reported methods on RoboCasa GR1 Tabletop~\citep{robocasa} and RoboTwin~2.0~\citep{robotwin2}.}
\label{tab:domain-generalization}
\footnotesize
\setlength{\tabcolsep}{1.5pt}
\begin{minipage}[t]{0.52\textwidth}
\centering
\textbf{(a) RoboCasa GR1 Tabletop}\par\smallskip
\begin{tabularx}{\linewidth}{@{}>{\raggedright\arraybackslash}Xr@{}}
\toprule
Method & Success (\%) \\
\midrule
\mbox{GR00T-N1.6~\citep{grootn16}} & 47.6 \\
DiT4DiT~\citep{dit4dit} & 50.8 \\
FLARE~\citep{flare} & 55.0 \\
LDA-1B~\citep{lda1b} & 55.4 \\
DIAL~\citep{dial} & 70.2 \\
WALA~\citep{wala} & 75.2 \\
LAWA~\citep{lawa} & 80.8 \\
\midrule
VAE-only & 78.8 \\
R\textsuperscript{2}-WAM & \textbf{84.7} (+5.9) \\
\bottomrule
\end{tabularx}
\end{minipage}\hfill
\begin{minipage}[t]{0.46\textwidth}
\centering
\textbf{(b) RoboTwin 2.0}\par\smallskip
\begin{tabularx}{\linewidth}{@{}>{\raggedright\arraybackslash}Xr@{}}
\toprule
Method & Avg. (\%) \\
\midrule
$\pi_{0.5}$~\citep{pi05} & 79.8 \\
Motus~\citep{motus} & 87.8 \\
StarVLA-$\alpha$~\citep{starvlaalpha} & 88.3 \\
InternVLA-A1~\citep{internvlaa1} & 89.5 \\
Fast-WAM~\citep{fastwam} & 91.8 \\
LingBot-VA~\citep{lingbotva} & 92.2 \\
WALA-Wan~\citep{wala} & 92.3 \\
\midrule
VAE-only & 90.6 \\
R\textsuperscript{2}-WAM & \textbf{92.8} (+2.2) \\
\bottomrule
\end{tabularx}
\end{minipage}
\end{table}

R\textsuperscript{2}-WAM achieves a state-of-the-art success rate of $84.7\%$ on RoboCasa GR1 Tabletop~\citep{robocasa}. On RoboTwin~2.0~\citep{robotwin2}, it reaches $92.8\%$. The gains over our VAE-only baselines are $5.9$ and $2.2$ percentage points, respectively, supporting the effectiveness of the same representation configuration across robot embodiments and environments.

\paragraph{Real-robot validation.}
Table~\ref{tab:r2wam-real-robot}(b) reports performance on the four physical tasks shown in Figure~\ref{fig:real-robot-tasks}. R\textsuperscript{2}-WAM improves complete-task success by $10.0$ percentage points on each of air-fryer food preparation, instructed pick-and-place, and cup pushing. On tool sweeping, the mean debris collection rate increases from $48.3\%$ to $66.7\%$. These results extend the improvements from the selected representation recipe to physical manipulation, including sequential manipulation and tool use.

\section{Conclusion}

In this work, we systematically study how the selection, integration, and composition of external visual representations affect VAE-based WAMs. Our findings suggest a practical recipe: start with a strong general-purpose representation, integrate it through channel-wise concatenation, and retain additional sources when they improve task success rates in the intended combination. R\textsuperscript{2}-WAM instantiates this recipe and demonstrates gains across the evaluated robot embodiments and environments.

Our findings provide a starting point for further research on visual representations and their use in WAMs. Future work should focus on training stronger general-purpose visual representations and developing integration mechanisms that improve future-feature prediction and enable action experts to use visual information more effectively.

\FloatBarrier

\bibliographystyle{iclr2027_conference}
\bibliography{references}

@article{fastwam,
  title={Fast-WAM: Do World Action Models Need Test-time Future Imagination?},
  author={Yuan, Tianyuan and Dong, Zibin and Liu, Yicheng and Zhao, Hang},
  journal={arXiv preprint arXiv:2603.16666},
  year={2026}
}

@article{wan,
  title={Wan: Open and Advanced Large-Scale Video Generative Models},
  author={{Wan Team} and Wang, Ang and Ai, Baole and Wen, Bin and Mao, Chaojie and Xie, Chen-Wei and Chen, Di and Yu, Feiwu and Zhao, Haiming and Yang, Jianxiao and others},
  journal={arXiv preprint arXiv:2503.20314},
  year={2025}
}

@inproceedings{motus,
  title={Motus: A Unified Latent Action World Model},
  author={Bi, Hongzhe and Tan, Hengkai and Xie, Shenghao and others},
  booktitle={Proceedings of the IEEE/CVF Conference on Computer Vision and Pattern Recognition},
  year={2026}
}

@article{dreamworld,
  title={DreamWorld: Unified World Modeling in Video Generation},
  author={Tan, Boming and Zhang, Xiangdong and Liao, Ning and Zhang, Yuqing and Zhang, Shaofeng and Yang, Xue and Fan, Qi and Zhang, Yanyong},
  journal={arXiv preprint arXiv:2603.00466},
  year={2026}
}

@inproceedings{repa,
  title={Representation Alignment for Generation: Training Diffusion Transformers Is Easier Than You Think},
  author={Yu, Sihyun and Kwak, Sangkyung and Jang, Huiwon and Jeong, Jongheon and Huang, Jonathan and Shin, Jinwoo and Xie, Saining},
  booktitle={International Conference on Learning Representations},
  year={2025}
}

@article{agra,
  title={Making Foresight Actionable: Repurposing Representation Alignment in World Action Models},
  author={Qiu, Lu and Li, Yizhuo and Chen, Yi and Ge, Yuying and Ge, Yixiao and Liu, Xihui},
  journal={arXiv preprint arXiv:2606.12217},
  year={2026}
}

@article{robustwam,
  title={Robust-WAM: Bridging Generative Pretraining and Semantic Foresight in World-Action Models},
  author={Yan, Haodong and Li, Junfeng and He, Junjie and others},
  journal={arXiv preprint arXiv:2608.05903},
  year={2026}
}

@article{stwam,
  title={ST-WAM: Semantic-Temporal World Action Model for Robust Manipulation under Visual Distribution Shifts},
  author={Wang, Mingxin and Hu, Bin and Qian, Bin and others},
  journal={arXiv preprint arXiv:2607.28993},
  year={2026}
}

@article{geosemwam,
  title={GeoSem-WAM: Geometry- and Semantic-Aware World Action Models},
  author={Ma, Fulong and Peng, Daojie and Yue, Wenjun and Cao, Jiahang and Wang, Bintao and Zhang, Qiang and Ma, Jun},
  journal={arXiv preprint arXiv:2606.03188},
  year={2026}
}

@article{repwam,
  title={RepWAM: World Action Modeling with Representation Visual-Action Tokenizers},
  author={Wang, Junke and Zhang, Qihang and Yang, Shuai and Luo, Yiming and Shen, Yujun and Wu, Zuxuan and Jiang, Yu-Gang and Xu, Yinghao},
  journal={arXiv preprint arXiv:2606.13674},
  year={2026}
}

@inproceedings{lda1b,
  title={LDA-1B: Scaling Latent Dynamics Action Model via Universal Embodied Data Ingestion},
  author={Lyu, Jiangran and Liu, Kai and Zhang, Xuheng and others},
  booktitle={Proceedings of Robotics: Science and Systems},
  year={2026}
}

@article{lawam,
  title={LaWAM: Latent World Action Models for Efficient Dynamics-Aware Robot Policies},
  author={Chen, Jialei and Wang, Kai and Chen, Kang and others},
  journal={arXiv preprint arXiv:2606.15768},
  year={2026}
}

@article{dinov3,
  title={DINOv3},
  author={Sim{\'e}oni, Oriane and Vo, Huy V. and Seitzer, Maximilian and others},
  journal={Transactions on Machine Learning Research},
  year={2026}
}

@inproceedings{vjepa21,
  title={{V-JEPA 2.1}: Unlocking Dense Features in Video Self-Supervised Learning},
  author={Mur-Labadia, Lorenzo and Muckley, Matthew and Bar, Amir and Assran, Mahmoud and Sinha, Koustuv and Rabbat, Michael and LeCun, Yann and Ballas, Nicolas and Bardes, Adrien},
  booktitle={European Conference on Computer Vision},
  year={2026}
}

@inproceedings{clip,
  title={Learning Transferable Visual Models From Natural Language Supervision},
  author={Radford, Alec and Kim, Jong Wook and Hallacy, Chris and others},
  booktitle={International Conference on Machine Learning},
  year={2021}
}

@article{siglip2,
  title={SigLIP 2: Multilingual Vision-Language Encoders with Improved Semantic Understanding, Localization, and Dense Features},
  author={Tschannen, Michael and Gritsenko, Alexey and Wang, Xiao and others},
  journal={arXiv preprint arXiv:2502.14786},
  year={2025}
}

@inproceedings{libero,
  title={LIBERO: Benchmarking Knowledge Transfer for Lifelong Robot Learning},
  author={Liu, Bo and Zhu, Yifeng and Gao, Chongkai and Feng, Yihao and Liu, Qiang and Zhu, Yuke and Stone, Peter},
  booktitle={Advances in Neural Information Processing Systems, Datasets and Benchmarks Track},
  year={2023}
}

@inproceedings{liberoplus,
  title={LIBERO-Plus: A Progressive Robustness Benchmark for Visual-Language-Action Models},
  author={Fei, Senyu and Wang, Siyin and Shi, Junhao and Dai, Zihao and Cai, Jikun and Qian, Pengfang and Ji, Li and He, Xinzhe and Zhang, Shiduo and Fei, Zhaoye and Fu, Jinlan and Gong, Jingjing and Qiu, Xipeng},
  booktitle={Proceedings of the IEEE/CVF Conference on Computer Vision and Pattern Recognition},
  year={2026}
}

@inproceedings{robocasa,
  title={RoboCasa: Large-Scale Simulation of Household Tasks for Generalist Robots},
  author={Nasiriany, Soroush and Maddukuri, Abhiram and Zhang, Lance and Parikh, Adeet and Lo, Aaron and Joshi, Abhishek and Mandlekar, Ajay and Zhu, Yuke},
  booktitle={Proceedings of Robotics: Science and Systems},
  year={2024}
}

@inproceedings{robotwin2,
  title={RoboTwin 2.0: A Scalable Data Generator and Benchmark with Strong Domain Randomization for Robust Bimanual Robotic Manipulation},
  author={Chen, Tianxing and Chen, Zanxin and Chen, Baijun and others},
  booktitle={International Conference on Machine Learning},
  year={2026}
}

@article{dreamzero,
  title={World Action Models are Zero-shot Policies},
  author={Ye, Seonghyeon and Ge, Yunhao and Zheng, Kaiyuan and Gao, Shenyuan and Yu, Sihyun and Kurian, George and Indupuru, Suneel and Tan, You Liang and Zhu, Chuning and Xiang, Jiannan and Malik, Ayaan and Lee, Kyungmin and Liang, William and Ranawaka, Nadun and Gu, Jiasheng and Xu, Yinzhen and Wang, Guanzhi and Hu, Fengyuan and Narayan, Avnish and Bjorck, Johan and Wang, Jing and Kim, Gwanghyun and Niu, Dantong and Zheng, Ruijie and Xie, Yuqi and Wu, Jimmy and Wang, Qi and Julian, Ryan and Xu, Danfei and Du, Yilun and Chebotar, Yevgen and Reed, Scott and Kautz, Jan and Zhu, Yuke and Fan, Linxi and Jang, Joel},
  journal={arXiv preprint arXiv:2602.15922},
  year={2026}
}

@article{lingbotva,
  title={Causal World Modeling for Robot Control},
  author={Li, Lin and Zhang, Qihang and Luo, Yiming and Yang, Shuai and Wang, Ruilin and Han, Fei and Yu, Mingrui and Gao, Zelin and Xue, Nan and Zhu, Xing and Shen, Yujun and Xu, Yinghao},
  journal={arXiv preprint arXiv:2601.21998},
  year={2026}
}

@article{lingbotva2,
  title={Native Video-Action Pretraining for Generalizable Robot Control},
  author={Zhang, Qihang and Li, Lin and Zhang, Luyao and Yang, Shuai and Luo, Yiming and Li, Shuaiting and Wang, Ruilin and Wang, Junke and Shao, Jiahao and Xu, Gangwei and Zhou, Jiaming and Shen, Yishu and Jin, Yudong and Xu, Fangyi and Ma, Shuailei and Liao, Jiaqi and Lu, Guanxing and Shi, Zifan and Wen, Yongkun and Zhao, Yujie and Tang, Weixuan and Wang, Xinyang and Li, Chaojian and Zhu, Jiapeng and Cheng, Ka Leong and Xue, Nan and Zhu, Xing and Shen, Yujun and Xu, Yinghao},
  journal={arXiv preprint arXiv:2607.08639},
  year={2026}
}

@article{flexpi,
  title={Flex-{$\pi$}: A Multi-Stream World-Action Model with Compute Flexibility},
  author={Yan, Ge and Liu, Jinghao and Fan, Yuzhi and Cai, Lei and Liao, Minwen and Zhang, Jesse and Fox, Dieter},
  journal={arXiv preprint arXiv:2608.10860},
  year={2026}
}

@inproceedings{vggtomega,
  title={{VGGT-$\Omega$}},
  author={Wang, Jianyuan and Chen, Minghao and Zhang, Shangzhan and Karaev, Nikita and Sch{\"o}nberger, Johannes and Labatut, Patrick and Bojanowski, Piotr and Novotny, David and Vedaldi, Andrea and Rupprecht, Christian},
  booktitle={Proceedings of the IEEE/CVF Conference on Computer Vision and Pattern Recognition},
  year={2026}
}

@article{sam3,
  title={{SAM 3}: Segment Anything with Concepts},
  author={Carion, Nicolas and Gustafson, Laura and Hu, Yuan-Ting and others},
  journal={arXiv preprint arXiv:2511.16719},
  year={2025}
}

@article{cradiov4,
  title={{C-RADIOv4} (Tech Report)},
  author={Ranzinger, Mike and Heinrich, Greg and McCarthy, Collin and Kautz, Jan and Tao, Andrew and Catanzaro, Bryan and Molchanov, Pavlo},
  journal={arXiv preprint arXiv:2601.17237},
  year={2026}
}

@inproceedings{flamingo,
  title     = {Flamingo: A Visual Language Model for Few-Shot Learning},
  author    = {Alayrac, Jean-Baptiste and Donahue, Jeff and Luc, Pauline and Miech, Antoine and others},
  booktitle = {Advances in Neural Information Processing Systems},
  year      = {2022}
}

@inproceedings{hrp,
  title={{HRP}: Human Affordances for Robotic Pre-Training},
  author={Srirama, Mohan Kumar and Dasari, Sudeep and Bahl, Shikhar and Gupta, Abhinav},
  booktitle={Proceedings of Robotics: Science and Systems},
  year={2024},
  address={Delft, Netherlands},
  doi={10.15607/RSS.2024.XX.068},
  url={https://www.roboticsproceedings.org/rss20/p068.html}
}

@article{dynaflip,
  title={{DynaFLIP}: Rethinking Robotics Perception via Tri-Modal-Dynamics Guided Representation},
  author={Lee, Jusuk and Lee, Seungjae and Shin, Jonghun and Jung, Hoseong and Kim, Sungha and Cho, Daesol and Kim, H. Jin and Huang, Jia-Bin and Huang, Furong},
  journal={arXiv preprint arXiv:2605.30350},
  year={2026},
  url={https://arxiv.org/abs/2605.30350}
}

@inproceedings{vc1,
  title={Where are we in the search for an Artificial Visual Cortex for Embodied Intelligence?},
  author={Majumdar, Arjun and Yadav, Karmesh and Arnaud, Sergio and Ma, Jason and Chen, Claire and Silwal, Sneha and Jain, Aryan and Berges, Vincent-Pierre and Wu, Tingfan and Vakil, Jay and Abbeel, Pieter and Malik, Jitendra and Batra, Dhruv and Lin, Yixin and Maksymets, Oleksandr and Rajeswaran, Aravind and Meier, Franziska},
  booktitle={Advances in Neural Information Processing Systems},
  volume={36},
  year={2023},
  doi={10.52202/075280-0031},
  url={https://proceedings.neurips.cc/paper_files/paper/2023/hash/022ca1bed6b574b962c48a2856eb207b-Abstract-Conference.html}
}

@inproceedings{unitok,
  title={{UniTok}: A Unified Tokenizer for Visual Generation and Understanding},
  author={Ma, Chuofan and Jiang, Yi and Wu, Junfeng and Yang, Jihan and Yu, Xin and Yuan, Zehuan and Peng, Bingyue and Qi, Xiaojuan},
  booktitle={Advances in Neural Information Processing Systems},
  volume={38},
  year={2025},
  doi={10.52202/085713-4306},
  url={https://proceedings.nips.cc/paper_files/paper/2025/hash/bbe04d329d531e8814f1199098bd8fb6-Abstract-Conference.html}
}

@inproceedings{videomaev2,
  title={{VideoMAE V2}: Scaling Video Masked Autoencoders with Dual Masking},
  author={Wang, Limin and Huang, Bingkun and Zhao, Zhiyu and Tong, Zhan and He, Yinan and Wang, Yi and Wang, Yali and Qiao, Yu},
  booktitle={Proceedings of the IEEE/CVF Conference on Computer Vision and Pattern Recognition},
  pages={14549--14560},
  year={2023},
  url={https://openaccess.thecvf.com/content/CVPR2023/html/Wang_VideoMAE_V2_Scaling_Video_Masked_Autoencoders_With_Dual_Masking_CVPR_2023_paper.html}
}

@misc{qwen35,
  title={{Qwen3.5}: Towards Native Multimodal Agents},
  author={{Qwen Team}},
  month={February},
  year={2026},
  url={https://qwen.ai/blog?id=qwen3.5}
}

@article{dial,
  title={{DIAL}: Decoupling Intent and Action via Latent World Modeling for End-to-End {VLA}},
  author={Chen, Yi and Ge, Yuying and Zhou, Hui and Ding, Mingyu and Ge, Yixiao and Liu, Xihui},
  journal={arXiv preprint arXiv:2603.29844},
  year={2026},
  url={https://arxiv.org/abs/2603.29844}
}

@article{lingbotvision,
  title={Vision Pretraining for Dense Spatial Perception},
  author={Fu, Zelin and Tan, Bin and Sun, Changjiang and Liu, Shaohui and Zheng, Kecheng and Xu, Yinghao and Zhu, Xing and Shen, Yujun and Xue, Nan},
  journal={arXiv preprint arXiv:2607.05247},
  year={2026},
  url={https://arxiv.org/abs/2607.05247}
}

@article{mecowam,
  title={Learning {4D} Geometric Priors for Inference-Efficient World Action Models},
  author={Zhang, Jianjun and Zhu, Jian and Su, Taiyi and Ma, Chong and Huang, Zitai and Xu, Yi and Wang, Hanli},
  journal={arXiv preprint arXiv:2607.05468},
  year={2026},
  url={https://arxiv.org/abs/2607.05468}
}

@article{fourdwam,
  title={{4D-WAM}: Infusing Spatiotemporal Awareness into World Action Models through Trajectory Fields},
  author={Yang, Lishan and Song, Wenxuan and Wang, Xi and Sheng, Pingyue and Fang, Zheng and Zhou, Ziyang and He, Junjie and Yan, Haodong and Chen, Jiayi and Sun, Nan and Sun, Qiao and Wang, Pengwei and Liu, Lingqiao and Wang, Yan and Gao, Yuxiang and Dayoub, Feras and Li, Haoang},
  journal={arXiv preprint arXiv:2608.08023},
  year={2026},
  url={https://arxiv.org/abs/2608.08023}
}

@article{wala,
  title={{WALA} Learning Executable Latent Actions from Action-Labeled Demonstrations and Action-Free Videos},
  author={Liu, Jiahao and Xia, Zhongpu and Tian, Shuai and Li, Huangrui and Zheng, Yuhang and Ma, Ning and Fu, Xin and Liu, Xiaotian and Li, Jing and Li, Yixian and Zhou, ShangQing and Xing, Zebin and Wang, Linbo and Li, Chaoyue and Li, Haoran and Zhao, Dongbin},
  journal={arXiv preprint arXiv:2607.11397},
  year={2026},
  url={https://arxiv.org/abs/2607.11397}
}

@misc{grootn16,
  title={{GR00T N1.6}: An Improved Open Foundation Model for Generalist Humanoid Robots},
  author={{NVIDIA GEAR Team}},
  year={2025},
  howpublished={NVIDIA Research},
  url={https://research.nvidia.com/labs/gear/gr00t-n1_6/}
}

@article{dit4dit,
  title={{DiT4DiT}: Jointly Modeling Video Dynamics and Actions for Generalizable Robot Control},
  author={Ma, Teli and Zheng, Jia and Wang, Zifan and Jiang, Chunli and Cui, Andy and Liang, Junwei and Yang, Shuo},
  journal={arXiv preprint arXiv:2603.10448},
  year={2026},
  url={https://arxiv.org/abs/2603.10448}
}

@article{flare,
  title={{FLARE}: Robot Learning with Implicit World Modeling},
  author={Zheng, Ruijie and Wang, Jing and Reed, Scott and others},
  journal={arXiv preprint arXiv:2505.15659},
  year={2025},
  url={https://arxiv.org/abs/2505.15659}
}

@inproceedings{pi05,
  title={{$\pi_{0.5}$}: A Vision-Language-Action Model with Open-World Generalization},
  author={Black, Kevin and Brown, Noah and Darpinian, James and others},
  booktitle={Proceedings of The 9th Conference on Robot Learning},
  pages={17--40},
  volume={305},
  series={Proceedings of Machine Learning Research},
  year={2025},
  url={https://proceedings.mlr.press/v305/black25a.html}
}

@article{starvlaalpha,
  title={{StarVLA-$\alpha$}: Reducing Complexity in Vision-Language-Action Systems},
  author={Ye, Jinhui and Gao, Ning and Yang, Senqiao and Zheng, Jinliang and Wang, Zixuan and Chen, Yuxin and Chen, Pengguang and Chen, Yilun and Liu, Shu and Jia, Jiaya},
  journal={arXiv preprint arXiv:2604.11757},
  year={2026},
  url={https://arxiv.org/abs/2604.11757}
}

@article{internvlaa1,
  title={{InternVLA-A1}: Unifying Understanding, Generation and Action for Robotic Manipulation},
  author={Cai, Junhao and Cai, Zetao and Cao, Jiafei and others},
  journal={arXiv preprint arXiv:2601.02456},
  year={2026},
  url={https://arxiv.org/abs/2601.02456}
}

@inproceedings{simonyan2013saliency,
  title={Deep Inside Convolutional Networks: Visualising Image Classification Models and Saliency Maps},
  author={Simonyan, Karen and Vedaldi, Andrea and Zisserman, Andrew},
  booktitle={International Conference on Learning Representations Workshop},
  year={2014}
}

@article{hoffman2019jacobian,
  title={Robust Learning with {Jacobian} Regularization},
  author={Hoffman, Judy and Roberts, Daniel A. and Yaida, Sho},
  journal={arXiv preprint arXiv:1908.02729},
  year={2019},
  url={https://arxiv.org/abs/1908.02729}
}

@article{yu2023sensory,
  title={Identifying important sensory feedback for learning locomotion skills},
  author={Yu, Wanming and Yang, Chuanyu and McGreavy, Christopher and Triantafyllidis, Eleftherios and Bellegarda, Guillaume and Shafiee, Milad and Ijspeert, Auke Jan and Li, Zhibin},
  journal={Nature Machine Intelligence},
  volume={5},
  pages={919--932},
  year={2023},
  doi={10.1038/s42256-023-00701-w},
  url={https://www.nature.com/articles/s42256-023-00701-w}
}

@article{t5,
  title={Exploring the Limits of Transfer Learning with a Unified Text-to-Text Transformer},
  author={Raffel, Colin and Shazeer, Noam and Roberts, Adam and Lee, Katherine and Narang, Sharan and Matena, Michael and Zhou, Yanqi and Li, Wei and Liu, Peter J.},
  journal={Journal of Machine Learning Research},
  year={2020}
}

@inproceedings{flowmatching,
  title={Flow Matching for Generative Modeling},
  author={Lipman, Yaron and Chen, Ricky T. Q. and Ben-Hamu, Heli and Nickel, Maximilian and Le, Matthew},
  booktitle={International Conference on Learning Representations},
  year={2023}
}

@article{lawa,
  title={Latent Action as Intention Enables Efficient Future Imagination for World Action Models},
  author={Li, Xiang and Zheng, Yupeng and Gu, Songen and Ma, Huailiang and Yu, Feng and Zheng, Yuhang and Nie, Xian and Yuan, Shanshuai and Zang, Yujie and Li, Weize and Tian, Shuai and Liu, Moyang and Zhang, Ya-Qin and Ding, Wenchao},
  journal={arXiv preprint arXiv:2608.24882},
  year={2026},
  url={https://arxiv.org/abs/2608.24882}
}

\clearpage
\appendix
\section{Visual Representation Construction}
\label{app:representation-construction}

\label{app:source-extension-slots}

\paragraph{Encoder configurations.}
We primarily use large (L) variants of the visual encoders. For Qwen3.5, we use Qwen3.5-4B.

\paragraph{Feature extraction.}
We use token features from the final layer of each encoder. For Qwen3.5, we extract the final-layer hidden states at visual-token positions from a single forward pass.

\section{Feature Integration Implementation Details}
\label{sec:feature-injection-implementation}

\subsection{WAM Formulations}
\label{app:wam-formulations}

\begin{figure}[!t]
\centering
\includegraphics[width=\linewidth]{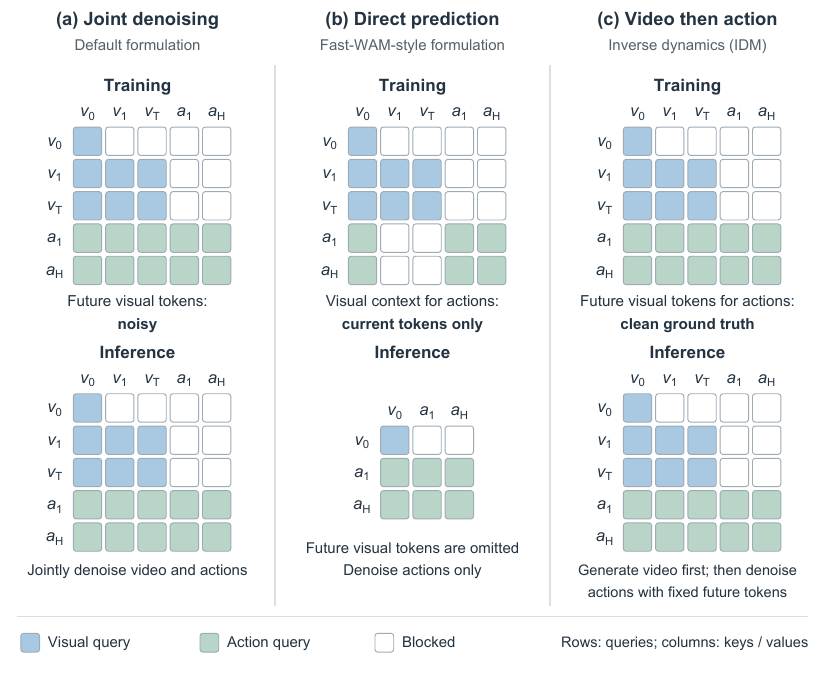}
\caption{Attention masks for training and inference in our three WAM formulations. Rows are queries and columns are keys/values; filled cells permit attention and empty cells block it. $v_0$, $v_{1:T}$, and $a_{1:H}$ denote current visual, future visual, and action tokens, respectively. Representative temporal positions are shown, with each visual entry representing a block of spatial tokens. Direct prediction omits future visual tokens at inference. Joint denoising and IDM share the displayed connectivity but differ in future-token conditioning: jointly denoised tokens versus clean ground-truth tokens during IDM training and previously generated, fixed tokens during IDM action inference.}
\label{fig:wam-attention-masks}
\end{figure}

We implement three WAM formulations within our base architecture, differing in the visual information available to the action expert and the training and inference procedures (Figure~\ref{fig:wam-attention-masks}).

\paragraph{Joint video--action denoising (default).}
Future visual tokens and action tokens are jointly denoised. As described in Section~\ref{sec:study-setting}, action tokens attend to current-frame visual tokens, noisy future visual tokens, and action tokens, whereas visual tokens do not attend to action tokens. At inference, future visual tokens and actions are initialized from noise and jointly denoised.

\paragraph{Direct action prediction.}
Following the direct-prediction paradigm of Fast-WAM~\citep{fastwam}, the attention mask restricts the action expert's visual context to current-frame tokens, preventing access to future visual tokens. At inference, actions are denoised conditioned on the current observation without generating future visual tokens.

\paragraph{Video-then-action prediction (IDM).}
Following the inverse-dynamics paradigm of LingBot-VA~\citep{lingbotva}, the action expert predicts actions conditioned on current and future visual tokens. During training, it receives clean ground-truth future visual tokens. At inference, the video expert first predicts future visual tokens, which then serve as conditions for action denoising.

\paragraph{VAE-only baselines.}
For each formulation, VAE-only denotes the corresponding model using VAE representations without additional external visual representations.

\subsection{Integration Mechanisms}
We use the notation introduced in Section~\ref{sec:study-setting} and Section~\ref{sec:feature-injection}: current and future VAE latents are $z_0^{\mathrm{vae}}$ and $z_{1:T}^{\mathrm{vae}}$, and the corresponding external features are $f_0$ and $f_{1:T}$. Observation indices $t=0,1,\ldots,T$ are distinct from flow time $\tau$, which is appended as a separate subscript for noised future states. Each integration mechanism adapts source features to its required spatiotemporal layout.

\noindent\textbf{Representation alignment (REPA).}
The student representation is the output of the eighth video-expert block.
A trainable three-layer MLP $P$ maps this state to the source feature
dimension, with SiLU after the first two linear layers. The teacher
trajectory is aligned to the video-token grid: spatial resizing is
bilinear, the observed temporal slice is preserved, and the eight future
slices are averaged into two groups of four. Padding masks are aligned with
the features and remove invalid locations. After $\ell_2$ normalization, the
alignment objective is
\begin{equation}
  \mathcal{L}_{\mathrm{REPA}}
  =
  1-\frac{1}{|\Omega|}\sum_{i\in\Omega}
  \left\langle
  \widehat{P(h_i^{8})},\widehat{f_i}
  \right\rangle,
\end{equation}
where $\Omega$ contains valid future tokens and excludes the observed slice.
The frozen teacher and the trainable projector are
used only during training and are absent from inference.

\noindent\textbf{Gated cross-attention.}
An independent feature cross-attention branch is inserted into each of the 30
action-expert blocks. The normalized action state after mixed attention is
used as the query, while normalized features from the current observation
$f_0$ provide keys and values through an adapter to the action-expert width. The original
T5 language cross-attention is retained, and both residual branches are
computed from the same post-mixed-attention state $\bar h_a^\ell$:
\begin{equation}
  \widetilde h_a^\ell
  =
  \bar h_a^\ell
  +\operatorname{CA}_{\mathrm{text}}^\ell(\bar h_a^\ell,c)
  +\tanh(g_\ell)\,
  \operatorname{CA}_{\mathrm{feat}}^\ell(\bar h_a^\ell,f_0).
\end{equation}
The resulting state is subsequently processed by the block FFN. Every layer
has its own scalar gate, initialized with $g_\ell=0$, so the added branch is
initially neutral. This mechanism has no auxiliary feature loss. The frozen
visual encoder and feature cross-attention branches remain active at inference.

\noindent\textbf{Channel-wise concatenation.}
Following the channel-wise integration design of DreamWorld~\citep{dreamworld},
we align external visual features with the VAE latent grid and concatenate
them along the channel dimension. This preserves the video token grid and
attention sequence length. The input and output projections are expanded to
accommodate the added features while retaining the pretrained VAE weights.
Observed features remain clean, whereas future VAE latents and external
features are jointly noised and predicted. At inference, future features are
initialized from noise and participate in joint denoising; only the VAE
latents are decoded.

\noindent\textbf{Token-wise concatenation.}
Dense external features are flattened into a separate token sequence, mapped
from $C_f$ to the video-expert hidden width by an independent input projection,
and appended to the VAE tokens. Feature and VAE tokens share all
video-expert Transformer blocks, while a separate prediction head maps feature
hidden states back to the source dimension $C_f$. Features are flattened in temporal, row, and column order into observed tokens $F_0$ and future tokens $F_{1:T}$. LayerNorm
precedes both the input projection and the prediction head; their linear
projections use Xavier-uniform weights and zero biases. The combined token
ordering is $[V_0,V_{1:T},F_0,F_{1:T},A]$.

Feature tokens use the video expert's three-dimensional rotary positions, with
feature-grid coordinates mapped to the video-token coordinate system. The
attention mask allows observed video and feature tokens to interact only with
observed visual tokens. Future video and feature tokens jointly attend to both
observed and future video/feature streams. Action tokens attend to observed and noisy future video/feature tokens,
as well as action tokens. Video and feature tokens do not attend to action tokens.

During training, $F_0$ remains clean and $F_{1:T}$ receives Gaussian noise sampled
independently from the video noise. Feature and video flows use the same
sampled timestep and the same scheduler, with feature velocity target
$\epsilon_f-f_{1:T}$, where $\epsilon_f$ is the sampled feature noise. The scheduler-weighted MSE is evaluated over valid future
feature tokens only. At inference, $F_0$ is extracted from the observed views
and kept clean, while $F_{1:T}$ is initialized from Gaussian noise and jointly
denoised with video and actions. The observed prefix $F_0$ is restored after
every solver step, and the final predicted feature trajectory is discarded.

\noindent\textbf{Feature expert.}
This mechanism introduces an independent 30-block Transformer expert with
hidden width 1024, FFN width 4096, 24 attention heads, and head dimension 128.
Its input and output projections map between the source dimension $C_f$ and
the expert hidden width.

At every layer, the video, feature, and action experts independently form
their Q/K/V tensors. The tensors are concatenated for masked joint attention,
then split by modality before each expert applies its own attention output
projection, text cross-attention, normalization layers, and FFN. Joint
attention does not imply parameter sharing: the feature expert owns
independent Transformer parameters. Its attention mask follows the same
routing as token-wise concatenation: action queries attend to observed and noisy
future video/feature tokens and action tokens, while video and feature queries
cannot attend to action tokens. Feature noise is independent of video noise, while
the feature and video flows share the same timestep and scheduler. The
observed feature prefix remains clean, and only valid future features enter
the flow-matching loss. During
inference, future features are initialized from noise and participate in joint
denoising; the observed prefix is restored after every step, and the predicted
feature trajectory is discarded.

Unless otherwise specified, auxiliary feature-prediction losses use a weight of 1.0. The feature expert uses a feature-prediction loss weight of 0.1, and REPA uses an alignment weight of 0.1. Sensitivity experiments across five visual encoders compare REPA depths of 8, 15, and 22 at this weight, and weights of 0.1 and 0.5 at depth 8 (Table~\ref{tab:repa-sensitivity}). Increasing the alignment weight does not improve the mean LIBERO-Plus success rate, and the mean varies by less than 0.1 percentage points across the four configurations, although individual encoders exhibit different preferences.

We compare five feature-integration mechanisms using a common WAM backbone and training budget. The feature preprocessing and auxiliary-loss settings for each mechanism are specified above.

\begin{table}[!htbp]
  \centering
  \small
  \setlength{\tabcolsep}{3pt}
  \caption{Implementation summary of the five feature-injection mechanisms.}
  \label{tab:feature-injection-implementation}
  \begin{tabularx}{\linewidth}{@{}*{6}{>{\raggedright\arraybackslash}X}@{}}
    \hline
    \textbf{Mechanism} &
    \textbf{Representation interface} &
    \textbf{Future-feature prediction} &
    \textbf{Transformer parameters} &
    \textbf{Auxiliary objective} &
    \textbf{Feature pathway at inference} \\
    \hline
    REPA &
    hidden-state alignment &
    No &
    base WAM only &
    cosine alignment &
    removed \\
    Gated cross-attention &
    current feature memory &
    No &
    added cross-attention &
    none &
    retained \\
    Channel-wise concat. &
    feature channels &
    Yes &
    shared video expert &
    feature flow matching &
    retained during denoising \\
    Token-wise concat. &
    feature tokens &
    Yes &
    shared video Transformer &
    feature flow matching &
    retained during denoising \\
    Feature expert &
    feature tokens &
    Yes &
    independent expert &
    feature flow matching &
    retained during denoising \\
    \hline
  \end{tabularx}
\end{table}

\begin{table}[!htbp]
  \centering
  \small
  \caption{\textbf{REPA sensitivity to alignment depth and loss weight.} LIBERO-Plus success rates (\%) across five visual encoders.}
  \label{tab:repa-sensitivity}
  \setlength{\tabcolsep}{9pt}
  \begin{tabular}{lrrrr}
    \toprule
    Encoder & $d=8$ & $d=15$ & $d=22$ & $d=8$ \\
    & $\lambda=0.1$ & $\lambda=0.1$ & $\lambda=0.1$ & $\lambda=0.5$ \\
    \midrule
    CLIP & 77.51 & 76.56 & 75.98 & 77.47 \\
    DINOv3 & 78.05 & 77.36 & 77.86 & 78.02 \\
    SigLIP~2 & 76.28 & 78.57 & 77.25 & 76.10 \\
    VGGT-$\Omega$ & 76.43 & 76.00 & 76.85 & 78.23 \\
    V-JEPA~2.1 & 75.36 & 75.50 & 75.89 & 73.73 \\
    \midrule
    \textbf{Mean} & \textbf{76.73} & \textbf{76.80} & \textbf{76.77} & \textbf{76.71} \\
    \bottomrule
  \end{tabular}
\end{table}

\FloatBarrier

\section{Training and Evaluation Configurations}
\label{app:training-configuration}

\subsection{Simulation Training}
\label{app:simulation-training}

Table~\ref{tab:training-configuration} lists the common training settings, and Table~\ref{tab:benchmark-training-configuration} summarizes the benchmark-specific configurations. Within the main LIBERO comparisons (RQ1--RQ3), these settings are held fixed while the representation source, integration mechanism, or source set varies as specified in the main text.

\begin{table}[!ht]
\centering
\caption{\textbf{Common training settings for the simulation benchmarks.}}
\label{tab:training-configuration}
\vspace{5pt}
\normalsize
\setlength{\tabcolsep}{9pt}
\renewcommand{\arraystretch}{1.08}
\begin{tabular}{lc}
\toprule
Parameter & Value \\
\midrule
Optimizer & AdamW \\
Initial learning rate & $1\times10^{-4}$ \\
Minimum learning rate & $1\times10^{-6}$ \\
Learning-rate schedule & Cosine decay \\
Warmup & Linear, first $5\%$ of training \\
Weight decay & $0.01$ \\
Adam betas & $(0.9, 0.95)$ \\
Gradient clipping & $1.0$ (maximum norm) \\
Numerical precision & bfloat16 \\
Action loss weight & $1.0$ \\
\midrule
Action horizon & 32 steps \\
Video frames per view & 9 \\
Frame sampling offsets & $0,4,\ldots,32$ (33-step window) \\
\bottomrule
\end{tabular}
\end{table}

\begin{table}[!ht]
\centering
\caption{\textbf{Benchmark-specific training configurations.} Image sizes are per view.}
\label{tab:benchmark-training-configuration}
\vspace{5pt}
\small
\setlength{\tabcolsep}{9pt}
\renewcommand{\arraystretch}{1.08}
\begin{tabular*}{\textwidth}{@{\extracolsep{\fill}}lccc@{}}
\toprule
Parameter & LIBERO & RoboCasa & RoboTwin~2.0 \\
\midrule
Training data & \begin{tabular}[c]{@{}c@{}}LIBERO\\(four suites)\end{tabular} & GR1 Tabletop & \begin{tabular}[c]{@{}c@{}}Clean + randomized\\demonstrations\end{tabular} \\
\addlinespace[2pt]
Number of tasks & 40 & 24 & 50 \\
Dataset size (demos) & 1,712 & 24,000 & 27,500 \\
\midrule
RGB views & 2 & 1 (ego) & 3 \\
Image resolution & $224\times224$ & $224\times224$ & $240\times320$ \\
Action dimension & 7 & 47 & 14 \\
\midrule
Global batch size & 128 & 1024 & 1024 \\
Training epochs & 10 & 5 & 5 \\
\bottomrule
\end{tabular*}
\par\vspace{5pt}
\begin{minipage}{\textwidth}
\footnotesize
Dataset sizes are counted before the validation split. The processed LIBERO datasets used by our runs contain 1,712 episodes; the original four-suite corpus contains 2,000 demonstrations (500 per suite)~\citep{fastwam}. RoboCasa contains 1,000 demonstrations per task. RoboTwin combines 2,500 clean-scene and 25,000 randomized-scene demonstrations~\citep{fastwam}.
\end{minipage}
\end{table}

\subsection{Simulation Evaluation and Reporting}
\label{app:shared-experimental-setup}
\label{app:nosuffix-snapshot}

\paragraph{Benchmark coverage and aggregation.}
LIBERO uses 50 rollouts per task across 40 tasks, giving 2,000 rollouts per evaluation; we report the average across its four equally sized suites. A complete LIBERO-Plus evaluation contains 10,030 episodes across seven perturbation categories. Overall LP success is the total number of successes divided by the total number of evaluated episodes. RoboCasa GR1 Tabletop uses 50 trials per task across 24 tasks. RoboTwin~2.0 uses 100 trials per task across 50 tasks in each of the clean and randomized settings; its reported average weights these settings equally.

\paragraph{Instruction processing.}
\label{sec:nosuffix-protocol}
The official LIBERO-Plus task instructions include appended descriptions of the perturbation types. We remove these descriptions for evaluation, while retaining the rewritten instructions in the language perturbation category.

\paragraph{Evaluation seeds.}
Results use one trained checkpoint per configuration. The RQ1 summaries and RQ3 composition results report means and sample standard deviations across three evaluation seeds.

\subsection{Real-Robot Training and Evaluation}
\label{app:real-robot-evaluation}

\paragraph{Robot setup.}
We conduct all four real-robot tasks on the same Galbot G1 platform, equipped with two arms and grippers. The model takes images from one head-mounted camera and two wrist-mounted cameras as input. Actions consist of joint positions and gripper states.

\paragraph{Training configuration.}
We use 200 demonstrations each for air-fryer food preparation and instructed pick-and-place, and 100 each for tool sweeping and cup pushing. Real-robot models are trained for 10 epochs with a batch size of 512.

\paragraph{Evaluation protocol.}
We evaluate the VAE-only WAM and R\textsuperscript{2}-WAM on the four real-robot tasks in Table~\ref{tab:r2wam-real-robot}, with 20 trials per task for each model. Complete-task success is reported for air-fryer food preparation, instructed pick-and-place, and cup pushing; tool sweeping is evaluated using the mean debris collection rate, as defined below.

\paragraph{Complete-task success.}
For air-fryer food preparation, instructed pick-and-place, and cup pushing, we report the percentage of trials in which the full task is completed. Air-fryer food preparation requires opening the air fryer, placing the corn cob inside, and closing it; instructed pick-and-place requires selecting the instructed object and placing it at the specified location; cup pushing requires pushing the paper cup next to the tissue box.

\paragraph{Debris collection rate.}
Each sweeping trial starts with six pieces of debris. Let $c_i\in\{0,\ldots,6\}$ denote the number swept into the dustpan in trial $i$. We report the mean collection rate, $\frac{100}{20}\sum_{i=1}^{20}\frac{c_i}{6}$, as a percentage. This metric measures partial task completion rather than binary success, with each rollout treated as one evaluation trial.

\FloatBarrier

\clearpage
\section{Feature Integration Results and Diagnostics for RQ2}
\label{app:full-mechanism-results}

\subsection{Full Source--Mechanism Results}
\label{app:perturbation-results}

Table~\ref{tab:full-source-interface} reports LIBERO averages and LP perturbation-level and overall success rates for the thirteen main-study representations across all five integration mechanisms.

\begingroup
\footnotesize
\setlength{\tabcolsep}{2.6pt}
\renewcommand{\arraystretch}{1.0}
\setlength{\LTcapwidth}{\textwidth}
\begin{longtable}{@{}lrrrrrrrrr@{}}
\caption{Detailed LIBERO and LIBERO-Plus results for REPA, gated cross-attention, channel-wise concatenation, token-wise concatenation, and the feature expert.}
\label{tab:full-source-interface}
\label{tab:teacher-interface-perturbations}
\label{tab:source-extension-perturbations}
\label{tab:teacher-interface-perturbations-a}
\label{tab:teacher-interface-perturbations-b}
\label{tab:teacher-interface-perturbations-d}
\label{tab:teacher-interface-perturbations-e}
\\
\toprule
& LIBERO & \multicolumn{8}{c}{LIBERO-Plus} \\
\cmidrule(lr){2-2}\cmidrule(lr){3-10}
Source & Avg. & Camera & Robot & Lang. & Light & Bkg. & Noise & Layout & Total \\
\midrule
\endfirsthead
\caption[]{Detailed LIBERO and LIBERO-Plus results (continued).}\\
\toprule
& LIBERO & \multicolumn{8}{c}{LIBERO-Plus} \\
\cmidrule(lr){2-2}\cmidrule(lr){3-10}
Source & Avg. & Camera & Robot & Lang. & Light & Bkg. & Noise & Layout & Total \\
\midrule
\endhead
\midrule
\multicolumn{10}{r}{\textit{Continued on next page}}\\
\endfoot
\bottomrule
\endlastfoot
VAE-only reference & 97.9 & 51.2 & 77.8 & 91.1 & 94.3 & 62.0 & 66.6 & 82.4 & 74.7 \\
\midrule
\rowcolor{black!6}
\multicolumn{10}{@{}l}{\textbf{REPA}}\\*
\addlinespace[2pt]
DINOv3 & 98.4 & 59.0 & 75.5 & 92.2 & 96.2 & 67.8 & 75.1 & 83.0 & 78.1 \\
V-JEPA~2.1 & 97.9 & 52.7 & 76.3 & 91.0 & 92.6 & 64.9 & 71.6 & 80.9 & 75.4 \\
VC-1 & 97.5 & 53.7 & 75.5 & 90.6 & 95.7 & 60.5 & 69.6 & 81.1 & 74.9 \\
LingBot-Vision & 98.0 & 57.2 & 76.2 & 91.8 & 95.3 & 63.1 & 75.1 & 81.0 & 76.9 \\
VideoMAE V2 & 98.5 & 58.9 & 78.1 & 92.3 & 95.5 & 64.9 & 73.7 & 82.9 & 77.8 \\
SigLIP~2 & 98.4 & 53.6 & 76.3 & 91.4 & 97.4 & 65.4 & 71.0 & 82.2 & 76.3 \\
Qwen3.5 & 98.1 & 53.5 & 78.5 & 92.8 & 95.7 & 63.8 & 68.8 & 81.0 & 75.9 \\
SAM~3 & 97.8 & 54.7 & 78.3 & 91.9 & 94.7 & 63.0 & 68.8 & 79.5 & 75.5 \\
VGGT-$\Omega$ & 98.2 & 53.4 & 78.6 & 93.0 & 96.5 & 62.6 & 72.4 & 80.7 & 76.4 \\
HRP & 98.0 & 53.4 & 75.1 & 90.3 & 96.2 & 60.8 & 73.0 & 81.6 & 75.5 \\
DynaFLIP & 97.9 & 53.5 & 75.2 & 92.1 & 96.1 & 66.4 & 76.2 & 81.2 & 76.8 \\
UniTok & 98.2 & 57.3 & 75.6 & 91.6 & 94.9 & 62.4 & 73.8 & 80.1 & 76.3 \\
C-RADIOv4 & 97.9 & 53.0 & 75.9 & 90.8 & 95.9 & 59.4 & 67.7 & 80.1 & 74.4 \\
\midrule
\rowcolor{black!6}
\multicolumn{10}{@{}l}{\textbf{Gated cross-attention}}\\*
\addlinespace[2pt]
DINOv3 & 97.9 & 51.1 & 74.3 & 89.9 & 91.9 & 61.3 & 69.0 & 82.0 & 73.9 \\
V-JEPA~2.1 & 98.1 & 48.8 & 79.0 & 89.5 & 94.7 & 63.7 & 66.7 & 81.3 & 74.3 \\
VC-1 & 97.9 & 53.7 & 78.7 & 89.9 & 95.1 & 64.9 & 70.1 & 81.1 & 75.8 \\
LingBot-Vision & 97.8 & 49.1 & 75.9 & 92.0 & 96.7 & 63.1 & 67.8 & 81.3 & 74.6 \\
VideoMAE V2 & 98.3 & 53.0 & 77.9 & 90.6 & 94.6 & 63.8 & 70.9 & 82.6 & 75.8 \\
SigLIP~2 & 97.9 & 53.4 & 77.2 & 89.2 & 95.7 & 64.6 & 68.5 & 82.3 & 75.4 \\
Qwen3.5 & 98.1 & 49.0 & 79.4 & 92.1 & 90.4 & 61.3 & 71.1 & 81.6 & 74.8 \\
SAM~3 & 98.5 & 52.6 & 77.4 & 90.3 & 94.3 & 61.2 & 72.0 & 81.1 & 75.3 \\
VGGT-$\Omega$ & 98.6 & 50.6 & 77.9 & 92.9 & 96.8 & 61.2 & 67.1 & 82.0 & 75.1 \\
HRP & 98.2 & 51.2 & 76.1 & 89.9 & 92.1 & 59.8 & 67.6 & 80.5 & 73.6 \\
DynaFLIP & 98.3 & 52.5 & 75.3 & 87.6 & 95.3 & 63.9 & 71.1 & 79.9 & 74.6 \\
UniTok & 98.2 & 50.0 & 74.2 & 89.9 & 93.0 & 63.9 & 69.8 & 81.6 & 74.2 \\
C-RADIOv4 & 98.3 & 49.1 & 74.1 & 88.9 & 94.9 & 62.6 & 66.0 & 82.2 & 73.5 \\
\pagebreak
\midrule
\rowcolor{black!6}
\multicolumn{10}{@{}l}{\textbf{Channel-wise concatenation}}\\*
\addlinespace[2pt]
DINOv3 & 98.4 & 64.2 & 82.5 & 93.8 & 98.5 & 83.9 & 86.0 & 83.6 & 84.0 \\
V-JEPA~2.1 & 98.5 & 63.0 & 80.9 & 90.9 & 96.8 & 80.7 & 84.4 & 84.2 & 82.4 \\
VC-1 & 98.5 & 60.8 & 78.1 & 93.3 & 98.4 & 82.0 & 86.7 & 83.0 & 82.5 \\
LingBot-Vision & 98.1 & 62.6 & 79.6 & 93.0 & 97.8 & 81.2 & 87.7 & 82.9 & 83.0 \\
VideoMAE V2 & 98.7 & 53.6 & 81.4 & 91.3 & 95.4 & 68.0 & 73.1 & 82.7 & 77.5 \\
SigLIP~2 & 98.5 & 60.7 & 85.0 & 93.8 & 97.9 & 82.4 & 83.4 & 83.0 & 83.1 \\
Qwen3.5 & 98.5 & 59.7 & 84.5 & 93.2 & 98.0 & 80.0 & 78.7 & 82.5 & 81.6 \\
SAM~3 & 98.3 & 57.8 & 80.8 & 93.7 & 95.6 & 75.3 & 81.5 & 82.4 & 80.6 \\
VGGT-$\Omega$ & 98.4 & 62.4 & 78.8 & 92.7 & 95.8 & 75.7 & 79.6 & 82.5 & 80.6 \\
HRP & 98.6 & 52.6 & 77.8 & 91.9 & 96.9 & 82.2 & 81.2 & 82.4 & 79.8 \\
DynaFLIP & 98.0 & 55.2 & 83.1 & 92.3 & 93.9 & 69.9 & 77.5 & 82.1 & 78.8 \\
UniTok & 99.1 & 63.0 & 85.2 & 93.9 & 96.7 & 76.6 & 77.1 & 83.9 & 81.9 \\
C-RADIOv4 & 98.2 & 62.6 & 81.0 & 92.2 & 97.3 & 86.2 & 86.6 & 83.6 & 83.5 \\
\midrule
\rowcolor{black!6}
\multicolumn{10}{@{}l}{\textbf{Token-wise concatenation}}\\*
\addlinespace[2pt]
DINOv3 & 97.5 & 49.8 & 77.7 & 92.8 & 95.6 & 68.3 & 70.8 & 81.7 & 76.1 \\
V-JEPA~2.1 & 91.2 & 19.0 & 57.3 & 62.1 & 68.6 & 33.3 & 43.3 & 64.5 & 49.5 \\
VC-1 & 97.9 & 55.0 & 77.4 & 92.1 & 95.5 & 60.3 & 69.8 & 81.8 & 75.8 \\
LingBot-Vision & 98.1 & 52.3 & 77.1 & 87.3 & 95.4 & 68.6 & 74.2 & 81.1 & 76.0 \\
VideoMAE V2 & 97.9 & 52.7 & 77.0 & 91.3 & 93.3 & 65.9 & 73.5 & 81.1 & 76.1 \\
SigLIP~2 & 94.3 & 35.6 & 68.6 & 84.2 & 89.5 & 59.7 & 56.1 & 76.9 & 66.4 \\
Qwen3.5 & 98.2 & 49.5 & 79.4 & 90.3 & 96.5 & 62.6 & 72.5 & 80.9 & 75.5 \\
SAM~3 & 98.3 & 45.2 & 76.9 & 87.8 & 95.4 & 69.1 & 68.6 & 82.1 & 74.3 \\
VGGT-$\Omega$ & 96.7 & 48.0 & 73.8 & 88.8 & 89.7 & 74.0 & 70.5 & 81.3 & 74.4 \\
HRP & 97.9 & 54.2 & 75.2 & 89.6 & 94.2 & 66.2 & 69.7 & 82.1 & 75.4 \\
DynaFLIP & 97.5 & 54.1 & 74.0 & 86.3 & 94.1 & 60.5 & 71.0 & 80.2 & 74.0 \\
UniTok & 98.4 & 57.5 & 77.6 & 91.9 & 96.9 & 63.5 & 74.6 & 81.4 & 77.4 \\
C-RADIOv4 & 96.8 & 49.9 & 72.5 & 84.5 & 93.5 & 67.2 & 72.0 & 80.9 & 73.8 \\
\midrule
\rowcolor{black!6}
\multicolumn{10}{@{}l}{\textbf{Feature expert}}\\*
\addlinespace[2pt]
DINOv3 & 98.3 & 57.4 & 81.2 & 92.5 & 96.2 & 66.9 & 77.9 & 80.7 & 78.7 \\
V-JEPA~2.1 & 98.8 & 47.8 & 78.6 & 90.0 & 94.1 & 59.0 & 70.3 & 79.2 & 73.9 \\
VC-1 & 97.9 & 54.5 & 79.9 & 95.0 & 89.8 & 66.0 & 72.3 & 81.8 & 76.9 \\
LingBot-Vision & 98.2 & 54.8 & 78.6 & 95.7 & 91.6 & 61.6 & 76.1 & 80.4 & 77.0 \\
VideoMAE V2 & 97.7 & 54.8 & 79.6 & 89.7 & 96.5 & 69.9 & 78.1 & 82.2 & 78.2 \\
SigLIP~2 & 98.1 & 53.7 & 80.1 & 93.7 & 95.9 & 64.4 & 73.3 & 81.2 & 77.2 \\
Qwen3.5 & 97.0 & 45.9 & 77.3 & 91.0 & 90.1 & 60.7 & 69.2 & 81.0 & 73.3 \\
SAM~3 & 98.3 & 57.3 & 76.8 & 91.3 & 95.7 & 61.5 & 74.1 & 81.8 & 76.8 \\
VGGT-$\Omega$ & 97.9 & 51.6 & 77.5 & 89.7 & 88.3 & 63.6 & 69.0 & 81.5 & 74.2 \\
HRP & 98.6 & 55.4 & 80.7 & 92.9 & 96.5 & 65.1 & 76.7 & 82.4 & 78.3 \\
DynaFLIP & 97.8 & 52.1 & 80.0 & 91.7 & 93.6 & 65.2 & 75.0 & 79.5 & 76.4 \\
UniTok & 98.9 & 50.5 & 79.2 & 94.0 & 94.6 & 62.2 & 70.8 & 81.3 & 75.8 \\
C-RADIOv4 & 98.0 & 53.3 & 76.6 & 91.7 & 95.5 & 66.7 & 74.1 & 81.9 & 76.7 \\
\end{longtable}
\noindent All values are success rates (\%). LIBERO reports the four-suite average; LP Total is the full-benchmark micro-average. Each mechanism is evaluated with the thirteen main-study sources.
\endgroup

\FloatBarrier

\subsection{Current-Feature Conditioning and Future-Feature Prediction}
\label{app:current-future-ablation}

We compare three DINOv3 configurations under channel-wise concatenation (Figure~\ref{fig:rq2_mechanism_analysis}(b)):

\begin{table}[!htbp]
\centering
\caption{DINOv3 prediction ablations under channel-wise concatenation.}
\label{tab:current-future-ablation}
\begin{tabular*}{\linewidth}{@{\extracolsep{\fill}}lccc@{}}
\toprule
Configuration & \shortstack{Current-feature\\input} & \shortstack{Future-feature\\prediction} & \shortstack{LP success\\(\%)} \\
\midrule
Current only & Yes & No & 76.5 \\
Future only & No & Yes & 81.9 \\
Current + future & Yes & Yes & 84.0 \\
\bottomrule
\end{tabular*}
\end{table}

The current-only configuration removes the future-feature channels and their prediction loss; no future features are generated at inference. The future-only configuration omits the current-frame DINOv3 features while retaining future-feature supervision and denoising. The full configuration includes both current-feature conditioning and future-feature prediction. All three configurations use the same attention-mask rule, training duration, remaining loss weights, and evaluation settings.

\paragraph{Feature prediction visualization.}
Following LDA-1B~\citep{lda1b}, we visualize DINOv3 features by projecting them onto three principal components and mapping these components to RGB colors. A shared PCA projection and color scale are used for the reference and predictions. The resulting feature maps are overlaid on the ground-truth future RGB image to illustrate their spatial structure.

\FloatBarrier

\subsection{Local Action Sensitivity Analysis}
\label{app:action-sensitivity}

We detail the gradient diagnostic in Figure~\ref{fig:rq2_mechanism_analysis}(d). The analysis is inspired by gradient-based sensitivity and attribution methods~\citep{simonyan2013saliency,hoffman2019jacobian,yu2023sensory}; our statistic uses local Jacobians.

\paragraph{Differentiation target and sampling path.}
Let $F$ denote the raw current-frame DINOv3 features from the agent-view and wrist cameras. Each view has a $14\times14$ grid of 1,024-dimensional features. For mechanism $m$, we write the final normalized action chunk as $A_m(F;\xi)\in\mathbb{R}^{32\times7}$, where $\xi$ collects the fixed non-feature inputs and realized sampling noise. We differentiate through the full ten-step denoising process, including each mechanism's transformations of the raw features, while holding the original VAE latents and other inputs fixed. The derivative therefore measures total local dependence on the current external-feature input across the sampling computation. It does not isolate dependence on a generated future-feature state.

We analyze channel-wise concatenation, token-wise concatenation, the feature expert, and gated cross-attention. REPA has no external-feature input at inference, so this diagnostic is not applicable to it. 

\paragraph{Action groups and gradient estimator.}
We flatten translation, rotation, and gripper outputs into separate vectors $a_g$ with dimensions $d_g=96$, $96$, and $32$, respectively. Within each camera, each non-overlapping $2\times2$ block of raw feature locations forms one cell $p$ on a $7\times7$ diagnostic grid. Let $F_{c,p}\in\mathbb{R}^{4096}$ contain all feature channels in this block, and let
\begin{equation}
J_{g,c,p}=\frac{\partial a_g}{\partial F_{c,p}}
\end{equation}
be its action--feature Jacobian. The local score is the Frobenius norm of this block Jacobian divided by $\sqrt{d_g}$, i.e., the root-mean-square input-gradient norm over the action outputs in group $g$.

To avoid constructing the full Jacobian, we use random output projections~\citep{hoffman2019jacobian}. For each group, draw independent Rademacher vectors $z_r\in\mathbb{R}^{d_g}$ with entries $\pm1/\sqrt{d_g}$ and compute vector--Jacobian products by automatic differentiation:
\begin{equation}
\widehat{s}_{g,c,p}
=\left[\frac{1}{R}\sum_{r=1}^{R}
\left\|\frac{\partial(z_r^{\mathsf T}a_g)}{\partial F_{c,p}}\right\|_2^2\right]^{1/2}.
\label{eq:action-sensitivity-estimator}
\end{equation}
Since $\mathbb{E}[z_rz_r^{\mathsf T}]=I/d_g$, the quantity inside the square root estimates $\|J_{g,c,p}\|_F^2/d_g$. We then average the local scores over the 49 cells and two camera views:
\begin{equation}
\widehat{S}_{m,g}(o)
=\frac{1}{2\cdot49}\sum_{c=1}^{2}\sum_{p=1}^{49}
\widehat{s}_{m,g,c,p}(o),
\label{eq:action-sensitivity-point}
\end{equation}
where $o$ identifies the observation state. Averaging occurs after the square root; this statistic is a mean of local sensitivity estimates, not the Frobenius norm of the entire camera Jacobian. 

\paragraph{Observation states and plotted points.}
All four mechanisms use the same nine fixed observation states: approach, grasp, and placement in each of three LIBERO trajectories. The recorded frame indices are $(28,36,93)$ for \texttt{spatial\_e000}, $(34,42,136)$ for \texttt{object\_e000}, and $(24,32,84)$ for \texttt{goal\_e114}, in approach--grasp--placement order. The \texttt{spatial\_e000\_f028\_approach} state uses $R=16$ probes per action group; the other eight states use $R=8$, consistently across mechanisms.

Each point in Figure~\ref{fig:rq2_mechanism_analysis}(d) is one $\widehat{S}_{m,g}(o)$ value. Colors distinguish translation, rotation, and gripper outputs, giving nine points per color and 27 points per mechanism. The horizontal axis is logarithmic; vertical offsets separate overlapping points and have no quantitative meaning. Points show variation across observation states, not independent training replications or confidence intervals.

\paragraph{Interpretation.}
For fixed other inputs and sampling noise, a small feature change satisfies $\Delta a_g\approx J_g\Delta F$ to first order. Larger scores therefore indicate stronger local action sensitivity in the measured feature coordinates.

\section{Feature Composition Protocol and Results for RQ3}
\label{app:composition-protocol}

\subsection{Feature Construction and Training Objective}

For source set $\mathcal K$, channel-wise concatenation uses aligned external features in a fixed source order. With $t=0$ indexing the current observation and $t=1,\ldots,T$ the future sequence, we write
\begin{equation}
\begin{aligned}
 f_t &= \operatorname{Concat}_{\mathrm{ch}}\{f_t^k:k\in\mathcal K\},\qquad t=0,1,\ldots,T,\\
 \widetilde z_0 &= [z_0^{\mathrm{vae}}\Vert_{\mathrm{ch}}f_0],\\
 \widetilde z_{1:T,\tau} &= [z_{1:T,\tau}^{\mathrm{vae}}\Vert_{\mathrm{ch}}f_{1:T,\tau}].
\end{aligned}
\label{eq:channel-concat}
\end{equation}
Here $f_{1:T,\tau}$ concatenates the source-wise future-feature states $f_{1:T,\tau}^k$ at flow time $\tau$ in the same order. The current state $\widetilde z_0$ remains clean. At inference, future VAE latents and external features are initialized from noise and jointly denoised. Each source contributes a future-feature loss:
\begin{equation}
\mathcal L_{\mathrm{aux}}=\sum_{k\in\mathcal K}\mathcal L_{\mathrm{aux}}^k,\qquad
\mathcal L_{\mathrm{aux}}^k=\mathcal L_{\mathrm{FM}}(f_{1:T}^k).
\label{eq:multi-feature-loss}
\end{equation}

The feature losses are evaluated on valid future targets $f_{1:T}^k$; current features $f_0^k$ remain clean and are excluded from the loss. Source-specific processing and concatenation order are held fixed between training and evaluation.

\subsection{Progressive Composition and Conditional Gains}

DINOv3 is the initial external source. The seven candidate additions are V-JEPA~2.1, SigLIP~2, SAM~3, VGGT-$\Omega$, HRP, UniTok, and C-RADIOv4. Figure~\ref{fig:progressive-composition} first evaluates each candidate added to DINOv3. The next group uses DINOv3 plus UniTok as its parent, and the four-source group uses DINOv3 plus UniTok plus V-JEPA~2.1. Each bar adds one source not already in its parent set. 

\subsection{Per-Seed Results and Evaluation Settings}

Table~\ref{tab:rq3-evaluation-seeds} reports the evaluations underlying the composition means and standard deviations. Each combination uses one fixed trained checkpoint, evaluated with seeds 41, 42, and 43. Every seed covers all 10,030 LP episodes. The mean is computed from the three full-benchmark success rates, and the standard deviation uses the sample convention. 

These composition evaluations use ten inference steps, replanning every ten steps, 30 initial waiting steps, no action ensembling, text guidance scale 1.0 and binarized gripper outputs. The LP evaluation protocol follows Appendix~\ref{app:nosuffix-snapshot}.

\begin{table}[!htbp]
\centering
\caption{RQ3 composition success rates on LIBERO-Plus across three evaluation seeds. All values are percentages; the standard deviation reflects evaluation-seed variability for a fixed trained checkpoint.}
\label{tab:rq3-evaluation-seeds}
\begingroup
\setlength{\tabcolsep}{7pt}
\renewcommand{\arraystretch}{1.05}
\begin{tabular*}{\linewidth}{@{\extracolsep{\fill}}lrrrr@{}}
\toprule
Source combination & Seed 41 & Seed 42 & Seed 43 & Mean $\pm$ SD \\
\midrule
$D+V$ & 85.13 & 84.62 & 85.03 & $84.93 \pm 0.28$ \\
$D+L$ & 86.01 & 85.50 & 85.97 & $85.83 \pm 0.28$ \\
$D+S$ & 84.93 & 83.80 & 84.75 & $84.49 \pm 0.61$ \\
$D+G$ & 84.38 & 84.29 & 84.55 & $84.40 \pm 0.13$ \\
$D+H$ & 85.31 & 83.91 & 84.23 & $84.48 \pm 0.74$ \\
$D+U$ & 86.05 & 85.85 & 86.47 & $86.12 \pm 0.32$ \\
$D+R$ & 79.74 & 77.82 & 78.23 & $78.59 \pm 1.01$ \\
\midrule
$D+U+V$ & 87.32 & 87.34 & 87.26 & $87.30 \pm 0.04$ \\
$D+U+L$ & 85.51 & 85.82 & 85.71 & $85.68 \pm 0.16$ \\
$D+U+S$ & 87.25 & 86.49 & 86.47 & $86.74 \pm 0.44$ \\
$D+U+G$ & 86.99 & 86.53 & 86.43 & $86.65 \pm 0.30$ \\
$D+U+H$ & 86.83 & 85.97 & 85.46 & $86.09 \pm 0.69$ \\
\midrule
$D+U+V+L$ & 87.60 & 87.40 & 87.51 & $87.50 \pm 0.10$ \\
$D+U+V+S$ & 85.84 & 86.38 & 86.30 & $86.17 \pm 0.29$ \\
$D+U+V+G$ & 85.62 & 84.83 & 84.36 & $84.94 \pm 0.64$ \\
$D+U+V+H$ & 87.56 & 87.29 & 88.33 & $87.73 \pm 0.54$ \\
\bottomrule
\end{tabular*}
\par\vspace{4pt}
\begin{minipage}{\linewidth}\small
$D$ = DINOv3, $V$ = V-JEPA~2.1, $L$ = SigLIP~2, $S$ = SAM~3, $G$ = VGGT-$\Omega$, $H$ = HRP, $U$ = UniTok, and $R$ = C-RADIOv4. The three groups correspond to the two-, three-, and four-source groups in Figure~\ref{fig:progressive-composition}.
\end{minipage}
\endgroup
\end{table}

\FloatBarrier

\section{Task-Level RoboCasa Results}
\label{app:robocasa-task-results}

\begingroup
\setlength{\LTcapwidth}{\textwidth}
\begin{longtable}{@{}lrlr@{}}
\caption{Task-level success rates (\%) of R\textsuperscript{2}-WAM on RoboCasa GR1 Tabletop. Each task is evaluated with 50 trials.}
\label{tab:robocasa-task-results}\\
\toprule
Task & \shortstack{Success\\(\%)} & Task & \shortstack{Success\\(\%)} \\
\midrule
\endfirsthead
\caption[]{Task-level success rates on RoboCasa GR1 Tabletop (continued).}\\
\toprule
Task & \shortstack{Success\\(\%)} & Task & \shortstack{Success\\(\%)} \\
\midrule
\endhead
\midrule
\multicolumn{4}{r}{\textit{Continued on next page}}\\
\endfoot
\bottomrule
\endlastfoot
Cup to drawer (close) & 54.0 & Placemat to bowl & 86.0 \\
Potato to microwave (close) & 86.0 & Placemat to plate & 88.0 \\
Milk to microwave (close) & 78.0 & Placemat to tiered shelf & 68.0 \\
Bottle to cabinet (close) & 88.0 & Plate to bowl & 90.0 \\
Wine to cabinet (close) & 76.0 & Plate to cardboard box & 76.0 \\
Can to drawer (close) & 96.0 & Plate to pan & 88.0 \\
Cutting board to basket & 94.0 & Plate to plate & 92.0 \\
Cutting board to cardboard box & 78.0 & Tray to cardboard box & 94.0 \\
Cutting board to pan & 98.0 & Tray to plate & 98.0 \\
Cutting board to pot & 88.0 & Tray to pot & 92.0 \\
Cutting board to tiered basket & 80.0 & Tray to tiered basket & 82.0 \\
Placemat to basket & 88.0 & Tray to tiered shelf & 74.0 \\
\midrule
\multicolumn{3}{l}{\textbf{Overall (1,016/1,200)}} & \textbf{84.7} \\
\end{longtable}
\endgroup
\FloatBarrier

\section{Task-Level RoboTwin Results}
\label{app:robotwin-task-results}

\begin{table}[H]
\centering
\caption{Task-level success rates (\%) of R\textsuperscript{2}-WAM on RoboTwin~2.0. Each of the 50 tasks is evaluated with 100 trials in each of the clean and randomized (Rand.) settings.}
\label{tab:robotwin-task-results}
\begin{tabular*}{\linewidth}{@{\extracolsep{\fill}}lrrlrr@{}}
\toprule
Task & Clean & Rand. & Task & Clean & Rand. \\
\midrule
Adjust bottle & 99.0 & 99.0 & Place container plate & 98.0 & 98.0 \\
Beat block hammer & 97.0 & 94.0 & Place dual shoes & 95.0 & 96.0 \\
Blocks ranking RGB & 100.0 & 97.0 & Place empty cup & 99.0 & 99.0 \\
Blocks ranking size & 66.0 & 87.0 & Place fan & 91.0 & 97.0 \\
Click alarmclock & 98.0 & 97.0 & Place burger fries & 100.0 & 99.0 \\
Click bell & 100.0 & 100.0 & Place mouse pad & 93.0 & 94.0 \\
Dump bin bigbin & 86.0 & 82.0 & Place object basket & 89.0 & 87.0 \\
Grab roller & 100.0 & 100.0 & Place object scale & 98.0 & 100.0 \\
Handover block & 99.0 & 96.0 & Place object stand & 99.0 & 99.0 \\
Handover mic & 100.0 & 100.0 & Place phone stand & 90.0 & 96.0 \\
Lift pot & 100.0 & 100.0 & Move pillbottle pad & 100.0 & 100.0 \\
Move can pot & 99.0 & 100.0 & Place shoe & 92.0 & 99.0 \\
Move playingcard away & 100.0 & 100.0 & Press stapler & 91.0 & 97.0 \\
Move stapler pad & 88.0 & 87.0 & Put bottles dustbin & 89.0 & 96.0 \\
Hanging mug & 79.0 & 75.0 & Put object cabinet & 89.0 & 81.0 \\
Open laptop & 95.0 & 98.0 & Rotate QR code & 88.0 & 81.0 \\
Open microwave & 56.0 & 45.0 & Scan object & 97.0 & 93.0 \\
Pick diverse bottles & 92.0 & 92.0 & Shake bottle & 100.0 & 99.0 \\
Pick dual bottles & 100.0 & 100.0 & Shake bottle horizontally & 100.0 & 99.0 \\
Place A2B left & 98.0 & 96.0 & Stack blocks three & 96.0 & 92.0 \\
Place A2B right & 97.0 & 98.0 & Stack blocks two & 100.0 & 100.0 \\
Place bread basket & 97.0 & 92.0 & Stack bowls three & 81.0 & 87.0 \\
Place bread skillet & 94.0 & 92.0 & Stack bowls two & 94.0 & 96.0 \\
Place can basket & 67.0 & 73.0 & Stamp seal & 94.0 & 98.0 \\
Place cans plasticbox & 100.0 & 100.0 & Turn switch & 57.0 & 66.0 \\
\midrule
\textbf{Overall} & \textbf{92.5} & \textbf{93.0} & \multicolumn{3}{r}{Combined: 92.8} \\
\bottomrule
\end{tabular*}
\end{table}
\FloatBarrier

\end{document}